\documentclass[11pt, a4paper, twocolumn]{berkeley-paul}
\usepackage{times}

\usepackage[numbers]{natbib}

\usepackage{nicefrac}       %
\usepackage{subcaption}
\hypersetup{
	colorlinks=true,
	linkcolor=blue,
	filecolor=magenta,      
	urlcolor=orange,
	citecolor=blue,
}
\usepackage{mathtools}
\usepackage{bm}
\usepackage{float}
\usepackage{siunitx}
\usepackage{wrapfig}
\usepackage{lipsum}
\usepackage{makecell}
\usepackage[capitalise, nameinlink]{cleveref}

\usepackage{footnote}
\makesavenoteenv{tabular}
\makesavenoteenv{table}
\usepackage{multirow}
\usepackage{booktabs}

\usepackage{algpseudocode}

\newcommand{\eg}{e.g., }
\newcommand{\Skip}[1]{}
\runningtitle{AutoEval: Autonomous Evaluation of Generalist Robot Manipulation Policies in the Real World}


\makeatletter
\renewcommand\Affilfont{\color{black}\normalfont\fontsize{10}{12}\selectfont}
\makeatother

\newcommand{\name}{AutoEval}
\newcommand{\website}{\url{https://auto-eval.github.io}}
\newcommand{\bigwebsite}{\scalebox{1.2}{\url{https://auto-eval.github.io}}}
\newcommand{\codesite}{\url{https://github.com/zhouzypaul/auto_eval}}

\begin{document}

\title{AutoEval: Autonomous Evaluation of Generalist Robot Manipulation Policies in the Real World}

\author{Zhiyuan Zhou$^{1}$, Pranav Atreya$^{1}$, You Liang Tan$^{1,2}$, Karl Pertsch$^{1}$, Sergey Levine$^{1}$ \\
$^{1}$UC Berkeley, $^{2}$NVIDIA \\
\bigwebsite
}

\teaserfigure{
    \vspace{-1.5em}
    \begin{center}
        \includegraphics[width=0.9\textwidth]{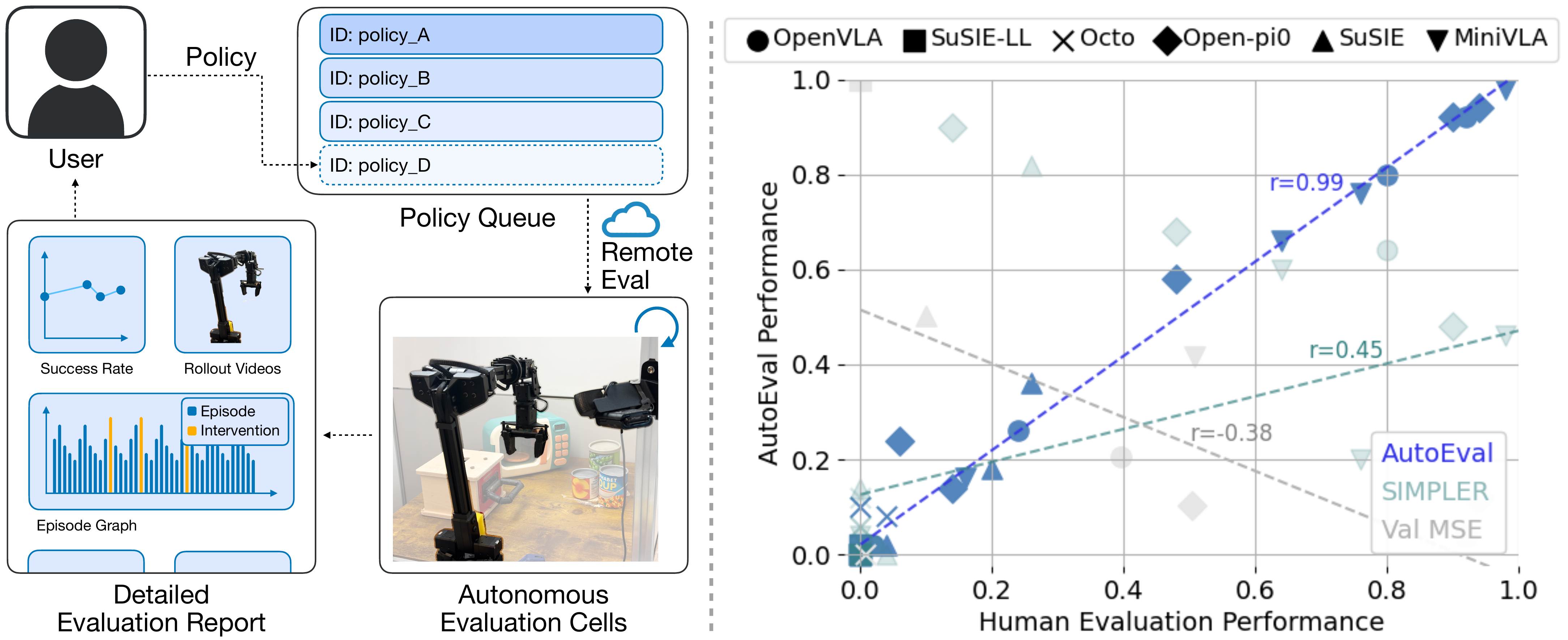}
        \captionof{figure}{We introduce \name, a system for scalable, automated real robot evaluation of generalist robot policies. %
        Automated evaluation results closely match human-run evaluations, while providing a more reliable performance signal than prior simulated evaluation approaches with photo-realistic environments (SIMPLER) or offline metrics such as validation error (Val MSE). \name{} reduces human supervision time for evaluation by more than 99\%. We provide public access to our AutoEval cells to facilitate standardization and ease of policy benchmarking.}
    \end{center}
    \label{fig:teaser}
}

\begin{abstract}
Scalable and reproducible policy evaluation has been a long-standing challenge in robot learning. Evaluations are critical to assess progress and build better policies, but evaluation in the real world, especially at a scale that would provide statistically reliable results, is costly in terms of human time and hard to obtain. Evaluation of increasingly generalist robot policies requires an increasingly diverse repertoire of evaluation environments, making the evaluation bottleneck even more pronounced. To make real-world evaluation of robotic policies more practical, we propose AutoEval, a system to autonomously evaluate generalist robot policies around the clock with minimal human intervention. Users interact with AutoEval by submitting evaluation jobs to the AutoEval queue, much like how software jobs are submitted with a cluster scheduling system, and AutoEval will schedule the policies for evaluation within a framework supplying automatic success detection and automatic scene resets. We show that AutoEval can nearly fully eliminate human involvement in the evaluation process, permitting around the clock evaluations, and the evaluation results correspond closely to ground truth evaluations conducted by hand. To facilitate the evaluation of generalist policies in the robotics community, we provide public access to multiple AutoEval scenes in the popular BridgeData robot setup with WidowX robot arms. In the future, we hope that AutoEval scenes can be set up across institutions to form a diverse and distributed evaluation network.
\end{abstract}

\maketitle

\vspace{-1em}
\section{Introduction}
Robot foundation models promise to drastically change the robot learning ``workflow'': instead of training policies for individual tasks or environments, these models are trained across a range of scenes, tasks, and robot embodiments~\citep{brohan2023rt, bousmalis2023robocat, octo_2023, kim2024openvla, shah2023gnm, sridhar2024nomad, doshi2024scaling, black2024pi_0, huang2023voxposer}, providing generalist policies that can solve new tasks in new settings. This shift to generalist training necessitates an analogous shift in how these policies are evaluated.
While traditional evaluations for single-task policies typically involve a few dozen policy rollouts that are practical to do by hand, robot foundation models may require hundreds of rollouts across a variety of tasks and scenes to obtain an accurate assessment of their generalist capabilities. For instance, a comprehensive evaluation of the recently introduced OpenVLA model ~\citep{kim2024openvla} against its baselines required more than 2,500 rollouts across four robot setups and three institutions, and a total of more than 100~hours of human labor for resetting scenes, rolling out policies, and recording success rates. Evaluations during the course of model development and design ablations may compound this effort multiple times over. Prior works have tried to address this evaluation bottleneck by building realistic simulated environments for evaluation~\citep{li24simpler}, but the gap between simulation and the real world can render results unreliable, and many tasks like cloth or liquid manipulation are challenging to simulate at sufficient fidelity. In this work we aim to develop a system for robot policy evaluation that combines the \emph{reliability} of real world evaluations, with the \emph{scalability} required for the evaluation of generalist robot policies. %

A key bottleneck for the scalability of real-world robot evaluations is the human operator time required to conduct the evaluation, reset the scene, and score policy success. If we can reduce required human involvement to a minimum, we can drastically increase the throughput of real robot evaluations by running evaluations around the clock. To this end, we propose \name, a system for designing \emph{autonomous} real-robot evaluations (see Figure 1). To use \name, human users queue policies for evaluation, which subsequently get evaluated with \emph{minimal} human intervention by the \name~system that automatically runs the policy, evaluates the results, resets the scene, and finally returns a detailed evaluation report to the user. 
\name~represents a new paradigm of real-world robot evaluation that has much higher throughput thanks to its minimal reliance on human intervention, allowing for much lower variance results with more trials per evaluation.

There are multiple challenges in designing an effective system for autonomous evaluation of real robot manipulation policies, such as the need for autonomous scene resets and success detection. Our work leverages large pre-trained models to \emph{learn} automatic reset policies and success detectors. Importantly, we adapt these models to the evaluation scene and task at hand to achieve high reliability and minimize the need for human intervention. We propose a general scheme for building automated robot evaluations and instantiate it for common tasks in the popular BridgeV2 robot evaluation environment~\citep{walke2023bridgedata}. 

Our central contribution is the development of an autonomous evaluation system, \name, that can evaluate user-supplied policies in the real world around the clock. We demonstrate that \name{} can scale to diverse evaluation environments by instantiating it in three automated evaluation environments for table-top manipulation tasks in the BridgeData~V2 environment~\citep{walke2023bridgedata}.
Our experiments show that the two aspects of evaluation that typically rely most on human effort, scene resets and success determination, can both be automated with high fidelity, yielding evaluation results that correlate well with ground truth human evaluations. \name~drastically increases the evaluation throughput, enabling 500~evaluation episodes per 24-hour period. We also find that \name{} provides a more reliable policy performance estimate than prior simulated evaluation approaches or offline metrics, while at the same time supporting a wider range of hard-to-simulate tasks like cloth manipulation.

We open-source our code~\footnote{\codesite} and a detailed step-by-step guide for setting up new \name{} platforms. Additionally, we open access to multiple Bridge-\name~cells, enabling researchers from other institutions to evaluate their policies on our Bridge-\name~systems. We hope that this takes a step towards democratizing robotics research and enabling fair comparisons of robot policies on unified evaluation setups.

\vspace{-0.5em}
\section{Related Work}

\textbf{Generalist robot policies. }
There has been significant progress in robot foundation models recently~\citep{brohan2023rt, kim2024openvla, octo_2023, kalashnikov2021mt, ehsani2023imitating, bharadhwaj2023roboagent, liu2024rdt, anil2023palm, sridhar2024nomad, ye2024latent, black2024pi_0, pertsch2025fast}, fueled by large-scale robot datasets~\citep{open_x_embodiment_rt_x_2023, walke2023bridgedata, khazatsky2024droid, shah2023gnm}. These models are trained to perform diverse tasks (e.g., pick-and-place, cloth folding)~\citep{walke2023bridgedata, kim2024openvla, black2024pi_0, pertsch2025fast}, adapt to various scenes with different backgrounds and distractors~\citep{zhou2024autonomous, fu2024context}, and control multiple robot embodiments (e.g., quadrupeds, manipulator arms, drones)~\citep{yang2024pushing, doshi2024scaling}. 
With the increase in capabilities of these generalist robot policies, evaluation becomes ever more time-consuming, because measuring model performance needs evaluations of a variety of different skills and scenes. For example, reporting results for \citet{kim2024openvla} required a few thousand evaluation trials and more than 100~hours of human labor. Evaluation trials needed during development probably compounded this number several times. This makes development and comprehensive evaluation of generalist robot policies increasingly challenging, calling for an evaluation method that is much more scalable.

\textbf{Robot policy evaluation in the real world. }
Evaluating robot policies in a fair, comprehensive, and reproducible way is challenging. Robotic methods and systems today are mostly tested in custom settings at the institution where the method is developed. Cross-institution evaluation encounters difficulties with different hardware, task definitions, and performance measures~\citep{van2018robotic}.
To address this, multiple works have proposed real robot setups that have reproducible components (such as 3-D printed objects or cheap robot hardware) that are meant to be replicated across institutions~\citep{yang2019replab, walke2023bridgedata, heo2023furniturebench, luo2023fmb, calli2015benchmarking, tyree20226, leitner2017acrv}. In addition to robot manipulators, there have also been efforts for standardized hardware in other robot embodiments~\citep{paull2017duckietown, pickem2017robotarium}. However, the sensitivity of policies to environmental factors like lighting, camera angles, and robot type makes it hard to accurately reproduce real robot setups across institutions, even when the same set of objects and hardware are used.
Others have built evaluation systems that are hosted at a central location to compare different approaches. Some take the form of live competitions~\citep{krotkov2018darpa, correll2016analysis, kitano1997robocup, van2018robotic, earthroverchallenge}, while others are hosted at research institutions and open to the public~\citep{zhou2023train, yenamandra2023homerobot}.
However, these evaluations all require human involvement to supervise the policy evaluation or to reset the scene, making it expensive in terms of human time and therefore significantly limiting the number of real robot evaluations benchmark participants can perform. In addition, the live competitions are logistically challenging and hard to operate continually.
These reproducibility and scalability constraints become even more apparent as the capabilities of robot policies expand to more scenes, tasks, and embodiments.
Our approach, \name, can substantially improve the throughput of real robot evaluations by replacing parts of the evaluation pipeline traditionally completed by humans with specialized learned components, thus enabling robots to ``evaluate themselves'' 24/7. 
Notably, \citet{bauer2022real} proposed a setup for remote, autonomous policy evaluation in the real world as part of their Real Robot Competition, but they focused on evaluations in a single environment, engineered to require no resets and allow for scoring with task-specific, hand-defined rules. In contrast, our \name{} system is designed for evaluation of \emph{generalist} policies by enabling autonomous evaluation on a wider range of tasks (e.g., pick-place, articulate object \& cloth manipulation) via learned reset and scoring modules.
While our goal is \emph{not} to build a comprehensive benchmark for robot foundation models, which requires evaluations spanning many tasks, scenes, and embodiments, we demonstrate that our system can be used to automate evaluations for a diverse set of tasks and provide a step-by-step guide to set up new automated evaluation within hours. We hope that by reproducing this recipe at other institutions, the robotics community will over time be able to construct a comprehensive evaluation benchmark for generalist policies.

\textbf{Evaluation in simulation.}
While human-run evaluations in the real world are the gold standard used by most prior works, they require extensive human effort and do not scale well as the capabilities of models increase. 
As a result, \emph{simulation} has been a popular tool for high-throughput evaluation in robot learning research~\citep{tassa2018deepmind, james2020rlbench, lee2021ikea, liu2024libero, nasiriany2024robocasa, mees2022calvin, makoviychuk2021isaac, tao2024maniskill3, kolve2017ai2, puig2023habitat, yu2020meta, ahmed2020causalworld, li2024behavior, li2021igibson, mandlekar2023mimicgen, mees2022matters}. 
However, there are still discrepancies between these simulators and the real world, making simulated evaluation different from real-world evaluation.
First of all, real-world physics of contacts, collisions, and friction are hard to simulate accurately~\citep{todorov2012mujoco, juliani2018unity, coumans2015bullet, lee2018dart, physx, xiang2020sapien, Genesis}. Even if the physics simulation is perfect, not all physical parameters can be precisely measured in the real world to replicate in simulation (for example, friction coefficients and actuation delays)~\citep{huang2023went, tobin2017domain}.
Policies that interact with real-world objects usually exhibit different behavior than they do on their simulated counterparts.
Secondly, policies need to deal with real world factors such as noisy and delayed sensory inputs that do not play a big part in simulation. 
Finally, the visual difference such as texture and lighting between simulated images and real-world observations makes the two types of evaluation quite different~\citep{deitke2020robothor, zhang2019vr}. Recent works have tried to reduce the visual discrepancy by building realistic simulators for policy evaluation~\citep{li2021openrooms, li24simpler}. SIMPLER~\citep{li24simpler} constructs high-fidelity replicas of real robot evaluation scenes and demonstrates strong correlation of simulated rollouts to human-run rollouts in the corresponding real robot environments. However, gaps between simulation and the real world remain, and our experiments show that they can affect different policies to varying degrees, leading to inconsistent policy performance rankings between simulation and real world evaluation. Additionally, a large number of tasks, like cloth or liquid manipulation, are challenging to simulate at sufficient fidelity to enable reliable evaluation.
In contrast, our approach performs evaluations on real robot systems and thus provides a more reliable signal for policy performance, including on tasks that are hard to simulate, while retaining scalability by minimizing the need for human intervention.

\textbf{Autonomous robot operations. }
Multiple prior works identified the need for human supervision as a key limiting factor in robot learning~\citep{zhou2024autonomous, gdm2024autort, kalashnikov2021mt, pinto2016supersizing, chen2021batch, lampe2023mastering}. While these works typically focus on autonomous policy \emph{improvement} instead of autonomous policy \emph{evaluation}, they share many challenges around robot resets and success detection. Thus, many of the techniques we employ for learning reset policies and success detectors are inspired by prior work in autonomous robot learning, and even some of the metrics are shared, \eg measuring the frequency of human intervention~\citep{beer2014toward}. However, to our knowledge, our work is the first to explore the design of a general system for autonomous evaluation of generalist policies.
While most robot learning researchers are (painfully) aware of the cost of evaluations, existing efforts toward automating real robot evaluations have been limited to task-specific solutions that often involve instrumenting the environment, \eg with spring-driven or scripted reset mechanisms~\citep{nagabandi2020deep, d2024achieving, kalashnikov2021mt}. In contrast, we provide a task-agnostic, scalable approach for automating robot evaluations with flexible, learned components based on generalizable and broadly applicable foundation models.

\vspace{-0.5em}
\section{Autonomous Evaluation of Robot Policies in the Real World}
\label{sec:method}

\begin{algorithm}[t]
\caption{Autonomous Policy Evaluation Loop}
\label{alg:autoeval}
\begin{algorithmic}[1]
\State \textbf{Input:} Task $T$, policy $\pi$ to be evaluated, initial state distribution $\rho(s)$, success classifier $\mathrm{C_T}$, reset policy $\pi_T$, reset classifier $\mathrm{C_{\rho(s)}}$
\State \textbf{Output:} Estimated prob. of success for task $T$
\For{each trial}
    \State \textbf{Start State:} Start from initial state $s_0 \sim \rho(s)$
    \State \textbf{Policy Rollout:} Rollout $\pi$ for $K$ steps
    \State \textbf{Success Check:} Label success using $\mathrm{C_T}(s_K)$
    \State \textbf{Reset Scene:} Rollout reset policy $\pi_T$ to return initial state to $\rho(s)$
    \State {\textbf{Failure:} If unable to reset or robot unhealthy, notify human operator to help}
\EndFor
\end{algorithmic}
\end{algorithm}

The policy evaluation problem setting we consider is rather straightforward: given a robot policy $\pi(a|o, l)$ that outputs actions given an observation $o$ and language instruction $l$, and a task definition $T:\mathrm{S} \rightarrow \{0,1\}$ that maps states to task success, we are interested in estimating the probability that the robot policy $\pi$ would be successful in completing the task $T$. The output of the policy evaluation is an evaluation score ranging from $0$ to $1$, representing the success probability.

During robot evaluations, the policy is typically asked to perform the same task multiple times, while applying randomizations to the initial state of the robot and the environment, to get a statistically significant estimate of the policy's performance under the initial state distribution $\rho(s)$. Conventionally, a human evaluator needs to be present for the full duration of the evaluation, supervising the robot, resetting the scene to a new initial position between trials, and scoring the policy's performance. Each individual trial may just take a few minutes, but for generalist policies that need to be evaluated across many tasks and trials, a comprehensive evaluation of a single checkpoint can quickly take multiple days. Thus, we next discuss our \name{} system for \emph{autonomous} policy evaluation that aims to minimize the required \emph{human} time for robot evaluation.

We present an overview of our \name{} system in \cref{alg:autoeval}. At its core, it follows the same structure as a conventional, human-run evaluation, running multiple trials with intermittent resets and performance scoring. However, \name{} introduces multiple learned modules that automatically perform the tasks that typically require a \emph{human} evaluator. Namely, \name{} consists of three key modules: (1)~a success classifier, that evaluates a policy's success on a given task, (2)~a reset policy, that resets the scene back to a state from the initial state distribution upon completion of a trial, and (3)~programmatic safety measures and fault detections that prevent robot damage and call for human intervention when necessary. All three components are implemented via flexible, \emph{learned} models, and can thus be easily adapted to automate the evaluation of a wide range of robot tasks. Next, we provide details on the design and training of each component of our \name{} system.

\textbf{Success classifier.} 
The success classifier $\mathrm{C_T}: \mathrm{S} \rightarrow \{0, 1\}$ serves to approximate the ground truth task-success $T: \mathrm{S} \rightarrow \{0, 1\}$ that maps image states to a binary success label. Instead of hand-crafting a task-specific success rule as done in prior work~\citep{gu2017deep, nagabandi2020deep}, \name{} \emph{trains} a learned success classifier $\mathrm{C_T}$, a recipe which can be easily applied to a wide range of robot tasks. Concretely, we collect a small set of example images of success and failure states. We use approximately $1000$ images, which takes less than $10$ minutes to collect by tele-operating the robot and saving the frames in the trajectory. We then fine-tune a pre-trained vision-language model (VLM) for the task of binary success detection. Given a language prompt, \eg ``Is the drawer open? Answer yes or no'', and an image observation, the model is trained to predict whether the task was successfully completed. We use a pre-trained VLM to obtain a classifier that is robust to small perturbations of the environment without needing to collect a large number of example images for fine-tuning. In practice, we use the Paligemma VLM~\citep{beyer2024paligemma} for training the success classifier, but many other open-source VLMs would be suitable.  More detailed information is provided in \cref{apdx:success-classifier}.  %

\textbf{Reset policy.} 
The reset policy $\pi_T(a|s)$ ``undoes'' what the evaluation policy $\pi$ did during the evaluation rollout, returning the scene and robot to a state from the initial state distribution $\rho(s)$. Again, instead of relying on task-specific ``hardware resets'' like springs or magnets, our aim with \name{} is to design a system that can be flexibly applied to a range of robot tasks. We thus use a \emph{learned} policy for resetting the scene. As we will show in \cref{sec:bridge-autoeval-details}, scripted reset policies can also be used in some tasks that have more structure, but learned policies provide a more generic approach that can be applied to a variety of tasks. To learn a reset policy, we manually collect a small set of approximately 100~high-quality demonstrations trajectories that reset the scene from plausible end-states of both successful and failed policy rollouts. In practice, this data collection takes typically less than two hours. We then fine-tune a generalist robot policy with behavioral cloning to act as a reset policy. Starting from a generalist policy checkpoint ensures that the reset policy is more robust, and fewer reset demonstrations are required to obtain reliable resets. %

\textbf{Safety detectors.} 
While success detector and reset policy \emph{in theory} enable autonomous evaluations, \emph{in practice} there are numerous issues and edge cases that can prevent evaluations from proceeding autonomously, like robot hardware failures, damage to scene or robot, or out-of-reach objects. In \name{} we use multiple measures to prevent or gracefully handle such issues. 
First, we implement a safety workspace boundary that the robot is constrained to, so policies with poor performance do not damage the robot or the \name{} scene.
Second, we implement programmatic checks of the robot's motor status and reboot motors if they failed e.g., due to a collision of the robot with the environment. We also train a ``reset success classifier'', similar to the success classifier above, that recognizes if resets were successful and re-runs the reset policy otherwise. In both cases, if multiple restarts or resets are not successful, e.g., because an object dropped from the workspace, we implement an automated notification system that requests manual intervention from an ``on-call'' human operator. In practice, our experiments show that such manual interventions are very rare for the \name{} cells we implemented (3~interventions per 24~hours of autonomous evaluation, see \cref{fig:long_study}).

\textbf{Setup time.} 
Overall, we find that the construction of an \name{} cell for a new task can be completed within 1-3h~of human effort, and less than 5~hours total, including model training time for success classifiers and reset policy. This is compared to tens of hours of human evaluation time that can be saved even within a single typical research project. We provide a detailed step-by-step guide for constructing new \name{} cells in \cref{sec:appx_step_by_step_guide} to make it easy for others to reproduce \name{} setups for their own tasks.

\vspace{-0.5em}
\section{Bridge-\name: Open-Source Automated Eval Platform}
\label{sec:bridge-autoeval-details}
\vspace{-0.2em}

\begin{figure}[t]
    \centering
    \includegraphics[width=\linewidth]{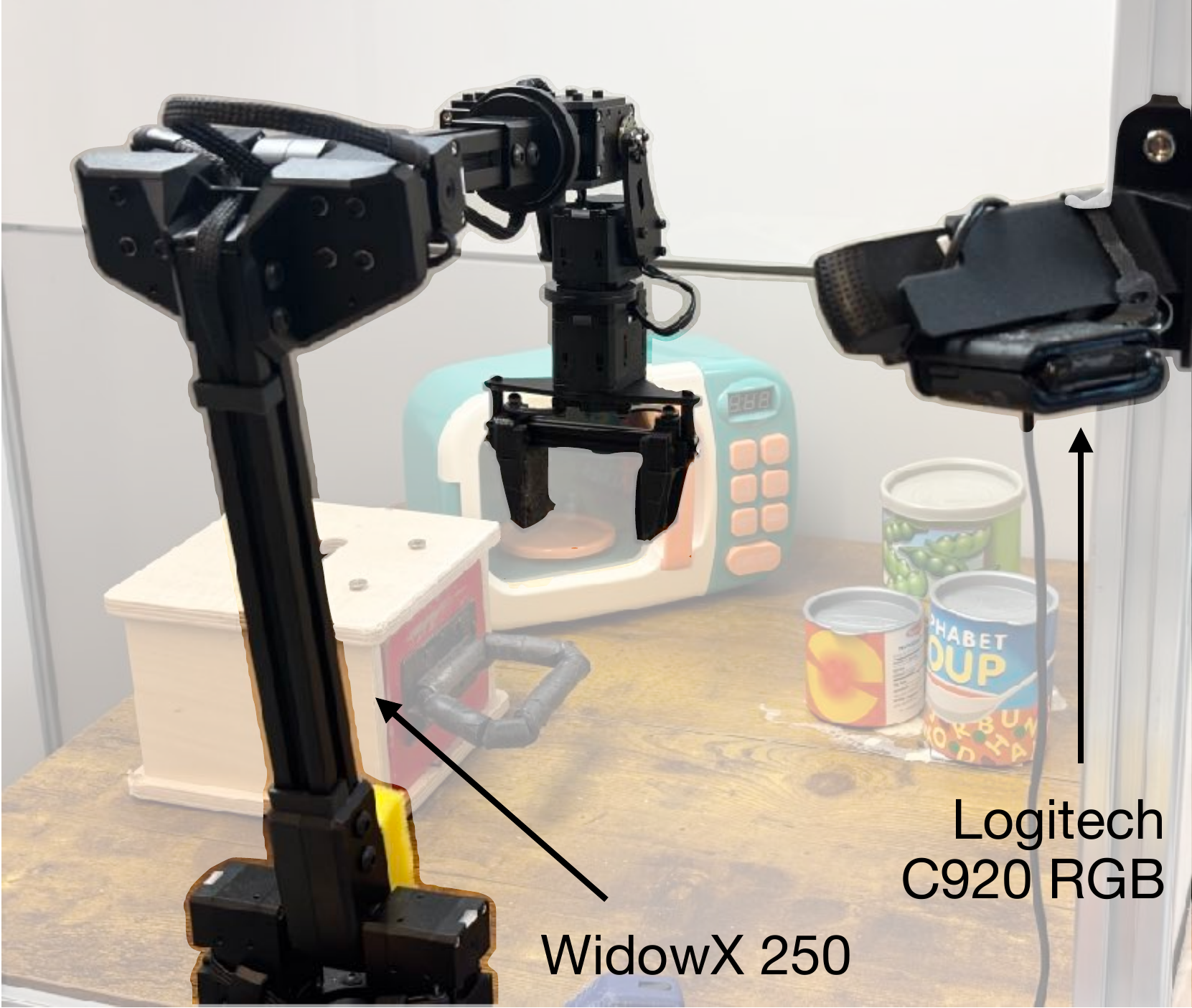}
    \caption{
        Bridge-\name{} cell: our robot setup for autonomous policy evaluation in the real world. It consists of a WidowX~250 6-DoF robot arm and a Logitech C920~HD RGB-camera. The scenes reproduce popular evaluation tasks from the BridgeData~\citep{walke2023bridgedata} robot dataset.
    }
    \label{fig:eval_cell}
    \vspace{-1em}
\end{figure}

In this section, we describe an instantiation of our automated evaluation system for multiple environments and tasks from the BridgeData~V2 dataset~\citep{walke2023bridgedata, ebert2021bridge}. BridgeData is a diverse manipulation dataset containing $60k$+ manipulation demonstrations with a WidowX 6DoF robot arm, that span $13$~different skills and $24$~environments. State-of-the-art generalist manipulation policies like OpenVLA~\citep{kim2024openvla}, RT2-X~\citep{brohan2023rt}, CrossFormer~\citep{doshi2024scaling}, and Pi0~\citep{black2024pi_0,pertsch2025fast} are all trained on BridgeData or a super set of it~\citep{open_x_embodiment_rt_x_2023},
and therefore policy evaluations on this setup are a natural testbed for scalable evaluation approaches for generalist policies.

\begin{figure}[t]
    \centering
    \includegraphics[width=0.32\linewidth]{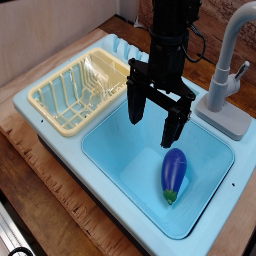}
    \includegraphics[width=0.32\linewidth]{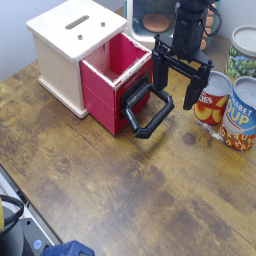}
    \includegraphics[width=0.32\linewidth]{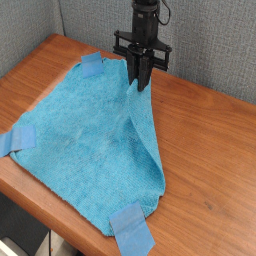}
    \caption{
        Three scenes in the Bridge-\name ~experiments: \texttt{sink}, \texttt{drawer}, and \texttt{cloth}. In total we support five tasks for autonomous evaluation: two pick-and-place task in \texttt{sink}, two drawer tasks in \texttt{drawer}, and one deformable cloth manipulation task in \texttt{cloth}.
    }
    \label{fig:tasks}
    \vspace{-1.5em}
\end{figure}

Similarly to \citet{walke2023bridgedata}, our Bridge-\name{} setup use a WidowX 250 6-DoF robot arm with a third-person Logitech C920~HD RGB-camera to capture the top-down $256 \times 256$ image of the robot workspace, as shown in~\cref{fig:eval_cell}. We use end effector delta action with blocking control. We built three Bridge-\name{} cells that can evaluate policies in parallel, as shown in \cref{fig:tasks}, which we call the \texttt{drawer} scene, the \texttt{sink} scene, and the \texttt{cloth} scene. We maintain constant lighting with an aluminum tripod light over each robot station. Each scene supports evaluation of one to two manipulation tasks: \texttt{drawer} supports evaluating ``open the drawer'' and ''close the drawer'';  \texttt{sink} supports evaluating pick-and-place tasks ``put the eggplant in the blue sink'' and ``put the eggplant in the yellow basket''; \texttt{cloth} support the deformable object manipulation task ``fold the cloth from top right to bottom left''. While none of the exact scenes are in the BridgeData dataset, all scenes are in the distribution of the tasks contained in BridgeData, and have been used in prior works to evaluate generalist policies~\citep{kim2024openvla, zhou2024autonomous, Zawalski24-ecot, black2023zero}. We choose these tasks since they represent diverse styles of manipulation tasks: pick-and-place, articulate object manipulation, and deformable object manipulation. %

For each scene, we train success classifiers and reset policies following~\cref{sec:method}. We also implement the safety detectors in \cref{sec:method} for the WidowX robot (see \cref{apdx:safety} for details), and an automated messaging system to request human interventions by sending push notifications programmatically to a Slack channel with ``on call'' operators for a given evaluation shift. 

\begin{figure}
    \centering
    \includegraphics[width=\linewidth]{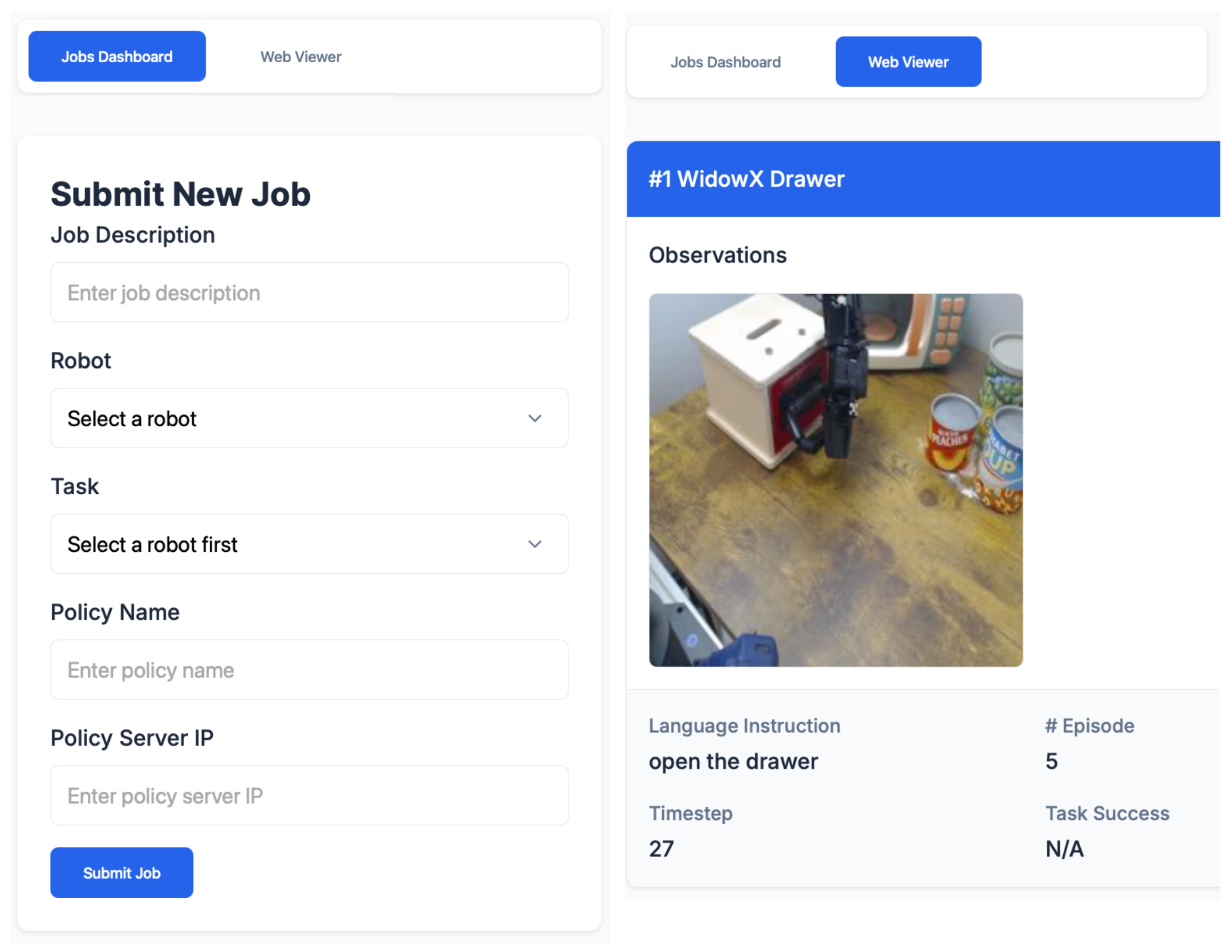}
    \caption{Web UI for submitting evaluation jobs to the Bridge-\name{} cells. Users choose a desired task and provide the IP address for a policy server they host for evaluation, and can monitor the evaluation through the UI.}
    \label{fig:submission-queue}
    \vspace{-1.5em}
\end{figure}

One contribution of our work is that we make two of our Bridge-\name{} cells \textbf{publicly available}, so other researchers can schedule evaluations for their policies. We hope that over time, this can contribute to making evaluations in robotics more reproducible and comparable. To make this practical, we provide a public web UI to access our Bridge-\name{} cells and monitor the evaluation progress, as shown in \cref{fig:submission-queue}.
Users can choose the task on which they want to perform evaluation, and provide the IP address for a ``policy server'', that serves the policy they want to evaluate. Given an image observations and a task instruction, the server runs the policy and returns a sequence of $7$D actions for the WidowX robot to execute (we provide example code for wrapping user policies in the server interface). 

\begin{figure}[t]
    \centering
    \includegraphics[width=\linewidth]{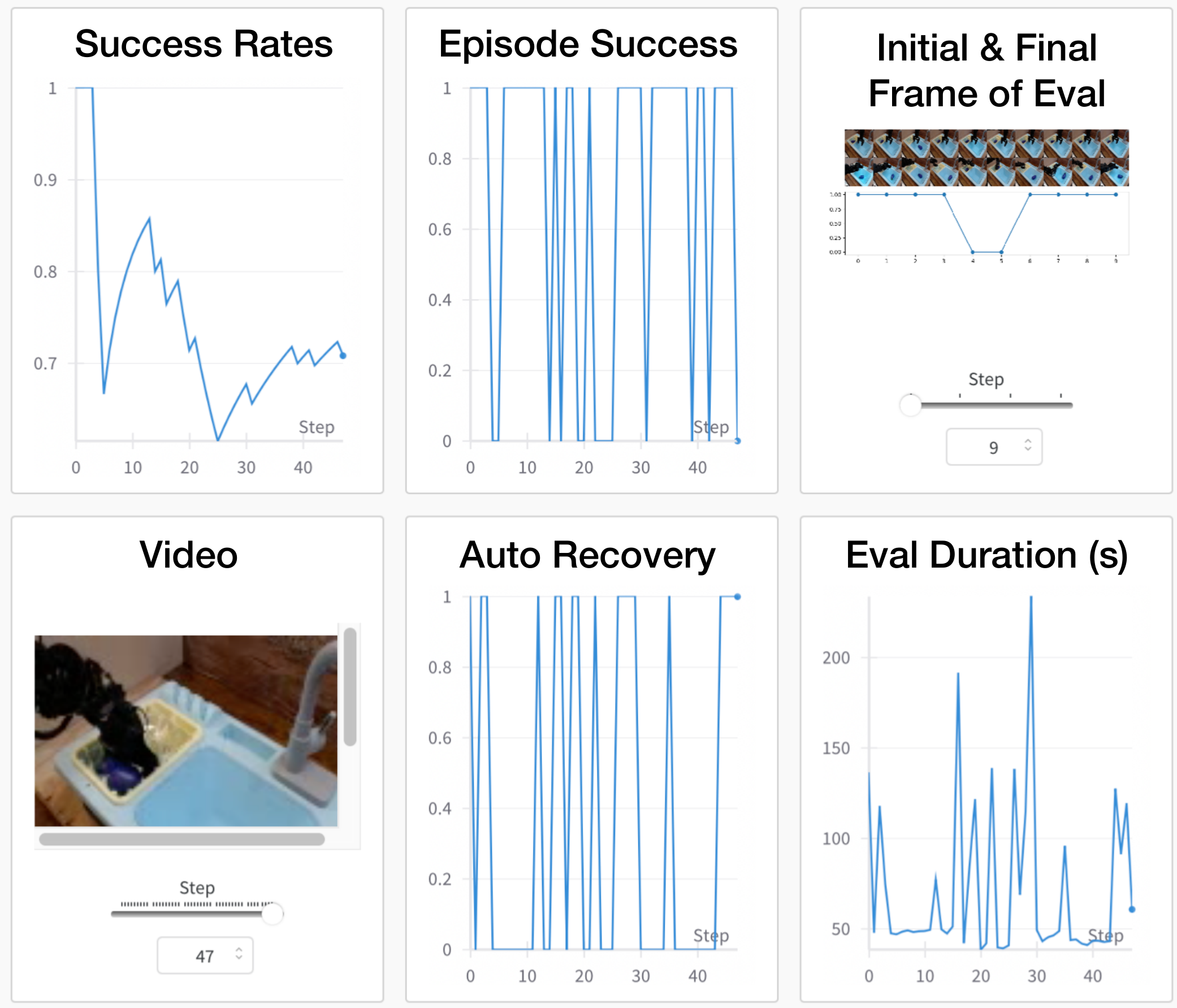}
    \caption{
        Excerpt from an \name{} result report, provided to the user upon completion of the automated evaluation. Users can see the per-episode success rate, rate of evaluation progress, instances of automatic recovery from motor failures, and qualitative rollout videos as well as classifier result plotted with initial and final frames to obtain a wholistic understanding of the policy's performance.
    }
    \label{fig:report-and-leaderboard}
    \vspace{-1em}
\end{figure}

Our Bridge-\name{} system will automatically queue the jobs for evaluation, and query the policy server for robot actions when the policy evaluation is executing. Our \name{} system can run around the clock, and execute evaluation jobs from all users in the order that they were submitted. At the end of a policy evaluation, \name{} provides users with downloadable rollout data and a detailed performance report of the autonomous evaluation, which contains rollout videos, success rates, episode durations, and frequencies of motor resets or required human interventions. \cref{fig:report-and-leaderboard} shows part of an example report, which is accessible online instantly after \name{} finishes. A step-by-step guide for submitting your policies to \name{} can be found at \website.

\vspace{-0.7em}
\section{Experimental Results}
\label{sec:results}

The goal of our experiments is to answer the following questions: (1)~How well does \name{}'s policy performance estimates match those of ``oracle'' human-run evaluations? (2)~Can \name{} evaluate policies more reliably and on a wider range of tasks than prior approaches for scalable evaluation of generalist policies? (3)~How stable is \name{} in operations over long periods of time and how effectively can \name{} minimize the amount of required human operator time?

\vspace{-1em}
\subsection{Experimental Setup}
\label{sec:experiment-setup}

\noindent \textbf{Tasks.} We evaluate policies on the five Bridge~V2~\citep{walke2023bridgedata} evaluation tasks described in \cref{sec:bridge-autoeval-details}: opening and closing a drawer, placing a plastic eggplant in a sink and a basket, and folding a piece of cloth. All tasks are performed using a WidowX 6-DoF robot arm. During human-run evaluations, success is counted when the drawer is completely closed or opened at least  1.5cm, respectively, if the eggplant is fully inside the sink or basket at the end of the episode, and if the cloth is folded to at least a quarter of the way diagonally. We randomize the initial position of the eggplant, drawer, and the cloth at the beginning of each episode. %

\noindent \textbf{Policies.} We run evaluations with six recently released generalist robot policies from the robotics community: \textbf{OpenVLA}~\citep{kim2024openvla}, a 7B parameter vision-language-action model (VLA) pre-trained on the Open X-Embodiment dataset~\citep{open_x_embodiment_rt_x_2023}, \textbf{Octo}~\citep{team2024octo}, a 27M parameter transformer policy, also pre-trained on Open X-Embodiment, \textbf{Open-$\bm{\pi_0}$}~\citep{openpizero}, an open-source reproduction of the 3B~parameter $\pi_0$ VLA~\citep{black2024pi_0} (the original $\pi_0$ was not available in open-source at the time of writing), pre-trained on the Bridge~V2 dataset, \textbf{MiniVLA}~\citep{belkhale2024minivla}, a 3B~parameter VLA pre-trained on the Bridge~V2 dataset~\citep{walke2023bridgedata}, \textbf{SuSIE}~\citep{black2023zero}, a hierarchical policy that combines a image diffusion subgoal predictor with a small diffusion low-level policy, pre-trained on Bridge~V2, and \textbf{SuSIE-LL}, which directly executes the goal-conditioned behavioral cloning low-level policy from SuSIE. This set of policies is a representative sample of current state-of-the-art generalist policies. All policies contain the Bridge~V2 dataset as part of their training data, and we evaluate the publicly released checkpoints for all models. 

\begin{figure}[t]
    \centering
    \includegraphics[width=0.48\linewidth]{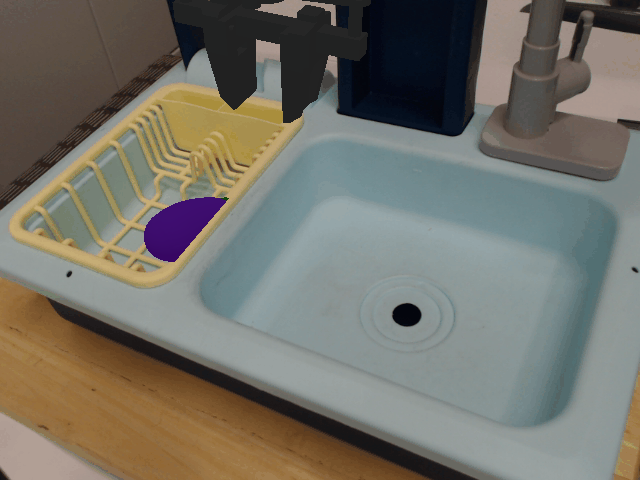}
    \includegraphics[width=0.48\linewidth]{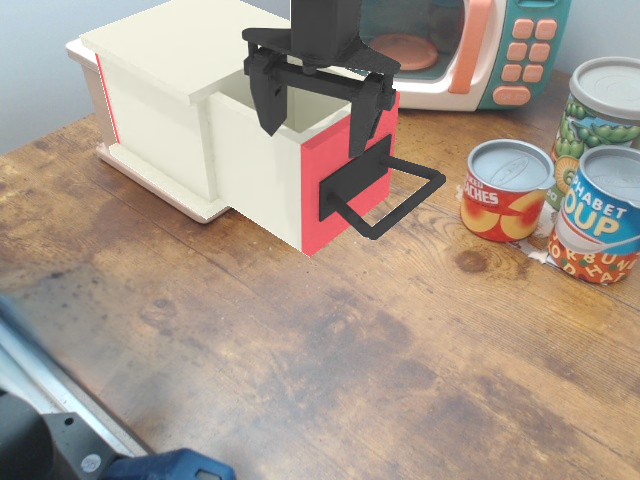}
    \caption{
        SIMPLER~\citep{li24simpler} simulated evaluation scenes for the tested environments. Simulated evaluation is fast and cheap, but can struggle from visual and physics discrepancies between simulation and the real world.
    }
    \label{fig:simpler-replication}
    \vspace{-1em}
\end{figure}

\noindent \textbf{Comparisons.} We compare multiple approaches for scalable evaluation of generalist policies. Concretely, we compare our approach, \textbf{\name{}}, to prior work on simulated evaluation of robot manipulation policies, \textbf{SIMPLER}~\citep{li24simpler}. SIMPLER builds realistic simulated versions of real-world environments and evaluates policies purely in simulation. For our experiments, we reuse the existing SIMPLER environment for the Bridge sink environment, and build a new SIMPLER simulation environment for the drawer scene (\cref{fig:simpler-replication}) by following \citet{li24simpler}'s step-by-step guide.
Deformable objects such as the cloth in our \texttt{cloth} scene are hard to simulate in general~\citep{ha2022flingbot, lin2021softgym}, and at the time of writing the simulator of SIMPLER, Maniskill~\citep{mu2021maniskill}, does not support simulating deformable objects so we do not evaluate the \texttt{cloth} scene in simulation.
In addition, we compare to using mean-squared error on a validation set (``\textbf{val-MSE}'') as a scalable approach for offline evaluation of robot policies.

\noindent \textbf{Metrics.} Human-run real-world evaluations represent a gold standard for robotic policy evaluation. For scalable evaluation approaches like the ones we compare in this work, the goal is to approximate the result of such human-run evaluations as closely as possible, while being significantly more scalable to run. Following \citet{li24simpler} we use two metrics to measure how closely the respective evaluation results match those of human-run evaluations: (1) \textbf{Pearson correlation}~\citep{pearson1895vii}, which measures the linear consistency between two random variables, and is a widely used statistical tool for assessing correlation, with scores nearing $1$ indicating high correlation. (2)~\textbf{MMRV} (Mean Maximum Rank Violation)~\citep{li24simpler}, which measures the consistency of policy \emph{ranking} and, as described in \citet{li24simpler}, can be more robust to noise on the evaluation results. 
MMRV is computed as follows: given $N$ policies $\pi_{1..N}$ and their respective success rates $R_{A, 1..N}$, $R_{B, 1..N}$ estimated via two evaluation procedures $A$ and $B$, we compute:
\begin{align*}
    \mathrm{RankViolation}(i, j) &= | R_{A, i} - R_{A, j} | \\
    &\quad\;\;\cdot \mathbf{1} [(R_{B, i} < R_{B, j}) \neq (R_{A, i} < R_{A, j})] \\
    \text{MMRV}(R_A, R_B) &= \frac{1}{N} \sum_{i=1}^N \max_{1 \leq j \leq N} \mathrm{RankViolation(i, j)}.
\end{align*}
For each evaluation approach we compute MMRV with reference to human-run ``oracle'' evaluations, where low MMRVs indicate closely matching evaluation results.

\subsection{\name{} Closely Matches Human Evaluation Results}
\label{sec:results-correlation}

\begin{figure}
    \centering
    \includegraphics[width=\linewidth]{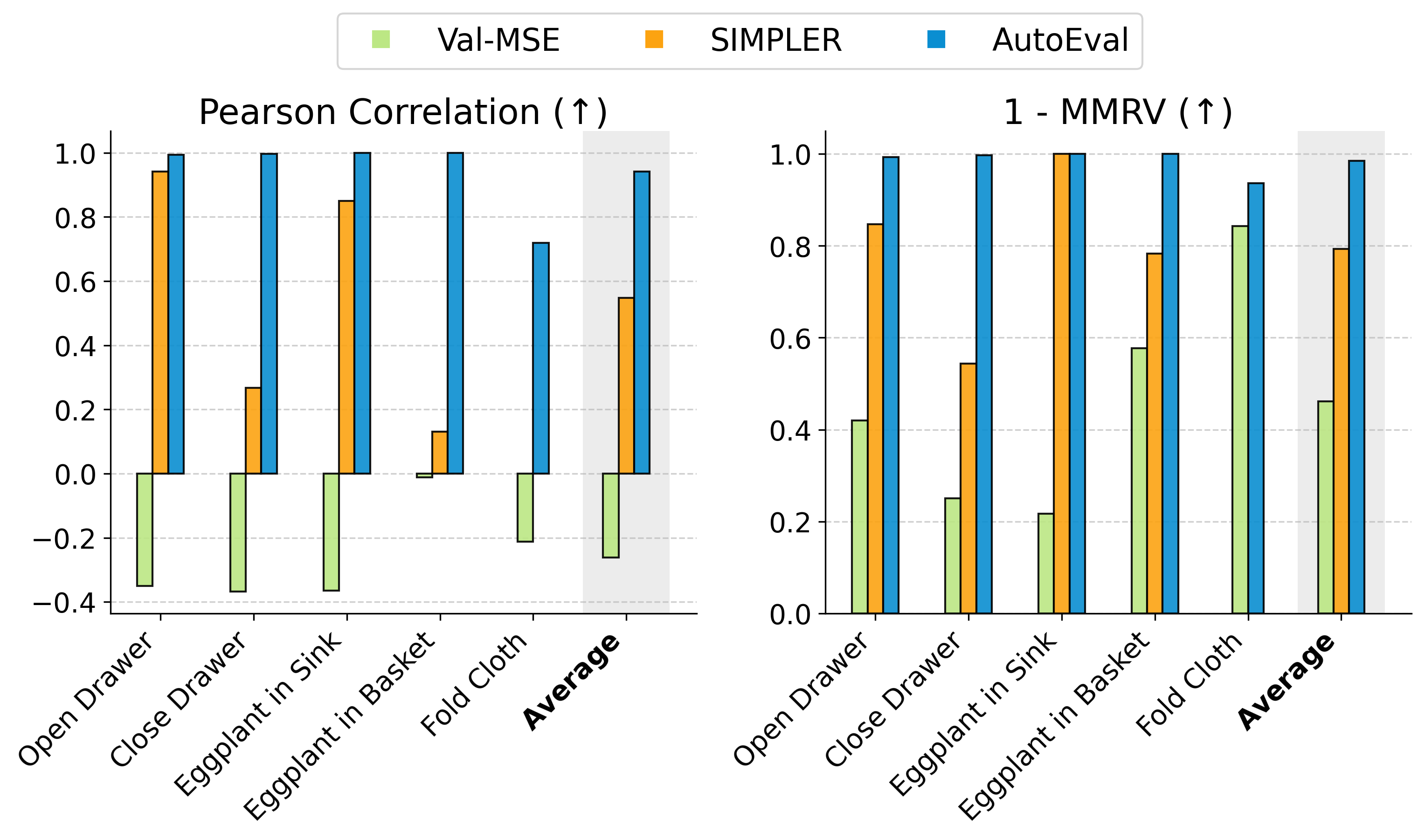}
    \caption{\textbf{Correlation of scalable evaluation approaches to oracle human-run evaluations.} \name{} closely matches human evaluations, achieving high correlation and low MMRV score (plotted in the figure is $1-\mathrm{MMRV}$ for clarity). In contrast, SIMPLER simulated evaluations and validation MSE do not correlate as well with human evaluations.}
    \label{fig:bar-correlation-plot}
    \vspace{-1em}
\end{figure}

In this section, we test how well the different evaluation approaches from \cref{sec:experiment-setup} match results from human-run evaluations. For each evaluation method, we run 50~evaluation rollouts for each policy in each task (except ``val-MSE'', which does not require rollouts).

We report results in \cref{fig:bar-correlation-plot}, with a detailed breakdown of results per task, policy, and evaluation method in Appendix, \cref{tab:autoeval-results} to \cref{tab:simpler-results}. Similar to prior work~\citep{li24simpler}, we find that simple validation MSE is a poor evaluation metric for robot policies: it actually negatively correlates with real robot performance and thus does not provide a reliable performance estimate. We find that SIMPLER evaluations in simulation provide a better performance signal, but lack reliability. Concretely, our results show that SIMPLER occasionally matches real-world performance well (e.g., for the ``open drawer'' task), but in other cases not accurately reflects the policy's performance. For example for Open-$\pi_0$ in the ``put eggplant to sink" task, the policy performs very poorly in simulated evaluations, but achieves high success rate in the real world. Intuitively, different policies may suffer differently from the remaining sim-to-real gap in SIMPLER evaluations. As a result, SIMPLER's effectiveness is policy dependent and it cannot provide a \emph{reliable} policy evaluation. 

\begin{figure*}[h]
    \centering
    \includegraphics[width=\linewidth]{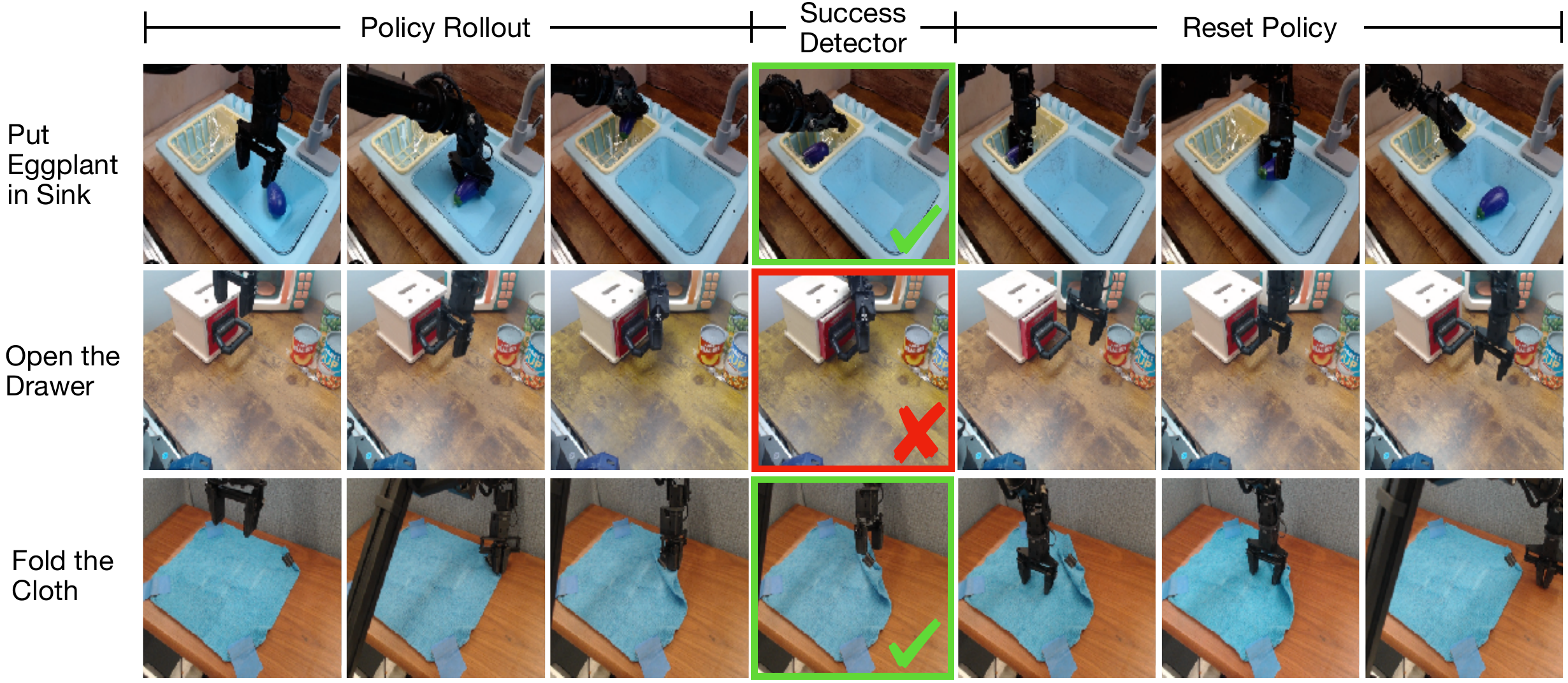}
    \caption{Qualitative visualization of \name{} evaluation rollouts on three of our tasks. After the policy execution is done, the success classifier determines whether the rollout was successful. Then, the reset policy returns the environment into a state from the initial state distribution for the next evaluation. Our evaluations cover representative robot manipulation tasks: pick-place, articulate and deformable object manipulation.}
    \label{fig:qualitative-results}
    \vspace{-1em}
\end{figure*}

In contrast, we find that our approach, \name{}, closely matches the results of oracle human-run evaluations, with an average Pearson score of $0.942$ and MMRV of $0.015$ (plotted as $1-\mathrm{MMRV}$ in Figure~\ref{fig:bar-correlation-plot} of $0.985$). In particular, an MMRV score close to zero indicates that it rarely disrupts the ranking of policies. Intuitively, since evaluations are still run in the real world, there is no sim-to-real gap that could negatively affect policy performance. In practice, we find that success detector and reset policy work reliably during evaluation. We show qualitative examples of autonomous evaluation rollouts in \cref{fig:qualitative-results}, and further examples in \cref{apdx:filmstrip-visualizations}. Importantly, we find that \name{} drastically reduces the human effort required to run real robot evaluations, cutting the human evaluator time for robot evaluations by $>99\%$ compared to conventional, human-run evaluations. We also note that \name{} does not perfectly match human-run evaluation results, due to occasional failures in success detection and reset policy. However, we find that in practice the accuracy of \name{} is sufficient to provide a strong ranking signal.

\vspace{-1em}
\subsection{\name{} Robustly Runs Over Long Time Spans}
\label{key:long_runtime_exp}

A key advantage of autonomous robot evaluations is that they can run 24/7, since they require little human involvement. In this section, we test \name{}'s stability when operating over extended periods of time, both in terms of its up-time and the reproducibility of policy evaluation performance.

\begin{figure*}[t]
    \centering
    \includegraphics[width=\linewidth]{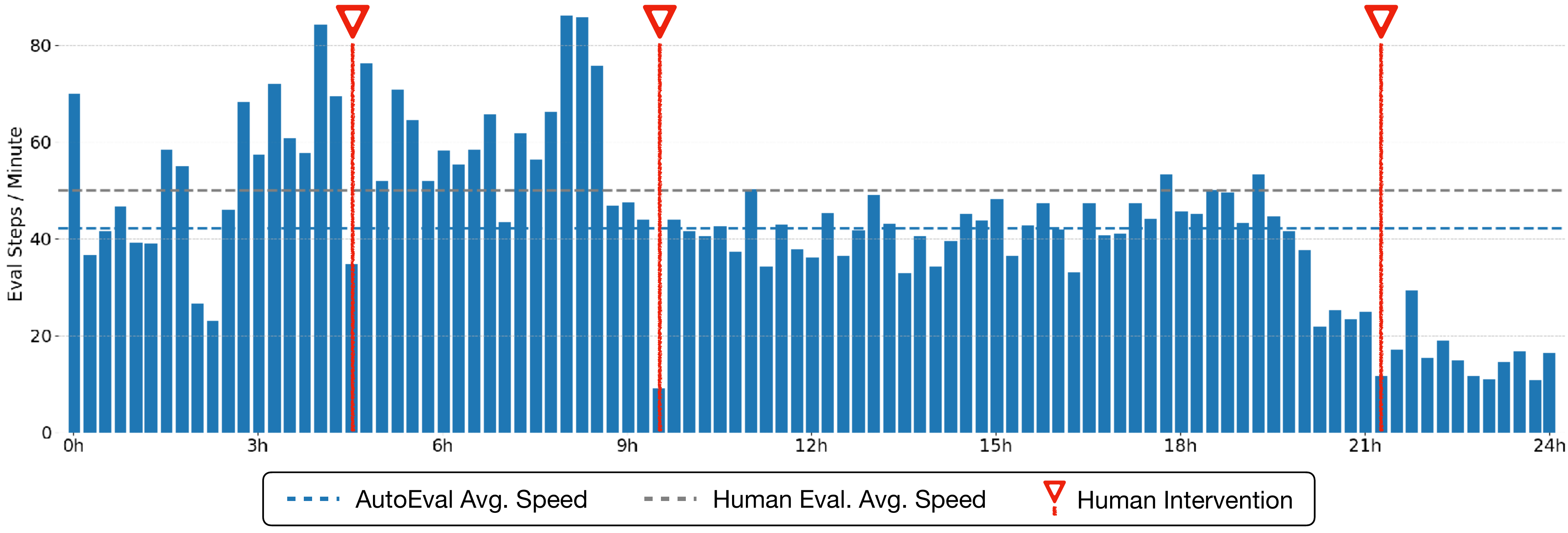}
    \caption{Visualization of a $24$~hour \name{} evaluation run with \textasciitilde 850~total evaluation episodes. \name{} ran autonomously over extended periods of time and only required a total of $3$~human interventions over a 24~hour period. On average, the evaluation throughput of \name{} is on par with that of human evaluations, but saves \textbf{99\%+} human operator time.}
    \label{fig:long_study}
    \vspace{-1em}
\end{figure*}

For this investigation, we performed a long-running evaluation over the course of 24~hours, repeatedly interleaving the evaluation of various policy checkpoints, using the ``\textit{open drawer}'' and ``\textit{close drawer}'' tasks.
In \cref{fig:long_study}, we present the evaluation throughput, as well as the number of human interventions needed over the span of the whole $24$ hours. We present evaluation throughput in terms of the number of valid evaluation steps taken per minute (excluding reset policy steps and re-evaluation steps needed because of motor failure). Over the course of a day, a single \name{} cell is able to run $60,000$ evaluation steps (roughly $850$ episodes on the drawer scene), with an average speed of $42$ evaluation steps per minute. The \name{} throughput varies in \cref{fig:long_study} because of the different inference speed of different policies.
The average \name{} speed, shown in dotted blue line, is slightly lower but on par with the average evaluation speed of a human evaluator performing manual resets of the environment and recording success rates.
Even though \name{} has a slightly lower throughput, \name{} runs autonomously and only required a total of three human interventions in the span of $24$ hours to reset the scene or robot. Every time a human operator needed to intervene, they simply needed to check and reset the objects' position in the scene, and potentially move the robot arm into reset position if a motors failed and the robot fell on the table. Afterward, the human operator can make \name{} resume simply with the press of a button. 
Assuming the average human response time during the day is $30$ minutes and $8$ hours at night, and that the $3$ required resets occur randomly throughout the $24$ hours, a whole day of \name{} yields $\approx19$ hours of ``real evaluation time'' that is not blocked by human reset. Assuming that each human reset operation takes $1$ minute, $19$h of real autonomous evaluation only costs $3$ minutes of human time, compared to $\approx16$ hours if a human evaluator wanted to run the same number of trials by hand. This means that \name{} can reduce human time involvement by \textbf{>99\%}.

\begin{figure}[h]
    \centering
    \includegraphics[width=\linewidth]{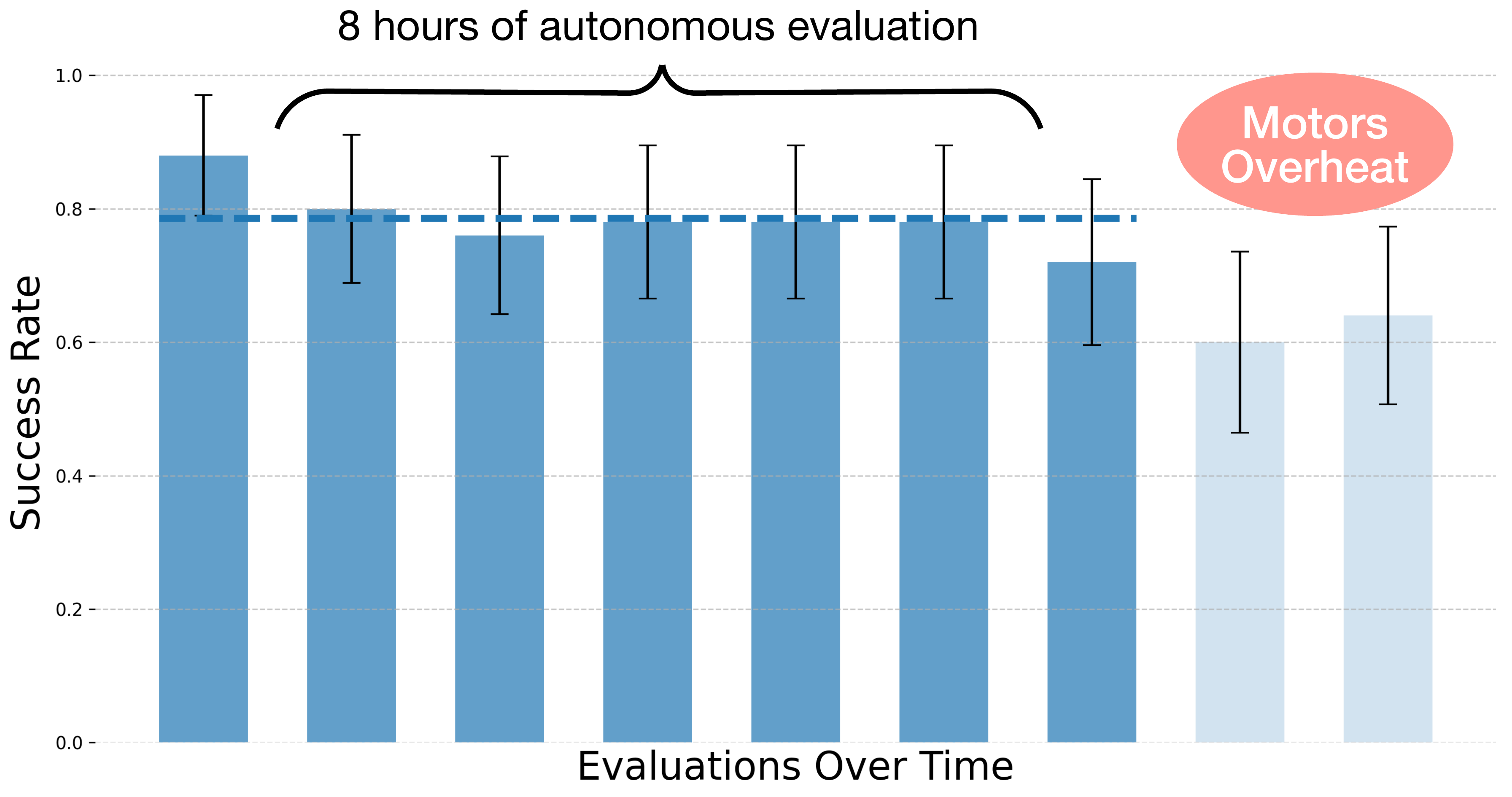}
    \caption{\name{} evaluation scores remain consistent over $8$ hours of autonomous evaluations. After 8~hours, WidowX motors overheat and evaluation scores start to drift. As a result, we pause evaluation for 20~min every 6~hours to let the motors cool off. Error bars show $95\%$ confidence intervals.}
    \label{fig:consistent-across-time}
    \vspace{-1.5em}
\end{figure}

\textbf{Are \name{} results consistent across time?}
We test the \emph{consistency} of \name{} evaluations, i.e., \name{}'s ability to produce comparable performance estimates across multiple iterations of evaluating the same policy. To test this, we run the Open-$\pi_0$ policy through a sequence of $9$ evaluations on the ``open drawer'' task, each consisting of $50$ individual trials, or a total of $450$ trials. Using \name{}, the full evaluation takes \textasciitilde $11$ hours. We report the results of this evaluation in \cref{fig:consistent-across-time}. We find that \name{} produces consistent evaluation results across long periods of time. Concretely, for the first $7$ evaluation runs, or a total of $350$ evaluation episodes, \name{} performance evaluation are within the margins of what might be considered the natural variance of robot evaluations ($\pm10\%$). We see a regression in performance after approximately $8$ hours of continuous operation, which we attribute to an overheating of the motors of our rather affordable WidowX robot (<\$3500) after many hours of operation. To mitigate the effects of such overheating in practice, we pause autonomous evaluations for 20~minutes every 6~hours to let the motors cool off before resuming evaluations.

In addition, we evaluate \name{}’s performance over two months of continuous operation. Results in \cref{apdx:two-months-reproducible} shows the \name{} yield consistent results over such long time periods.

\vspace{-0.5em}
\subsection{Analyzing \name{} Failure Modes}

\begin{figure}[h]
    \centering
    \includegraphics[width=\linewidth]{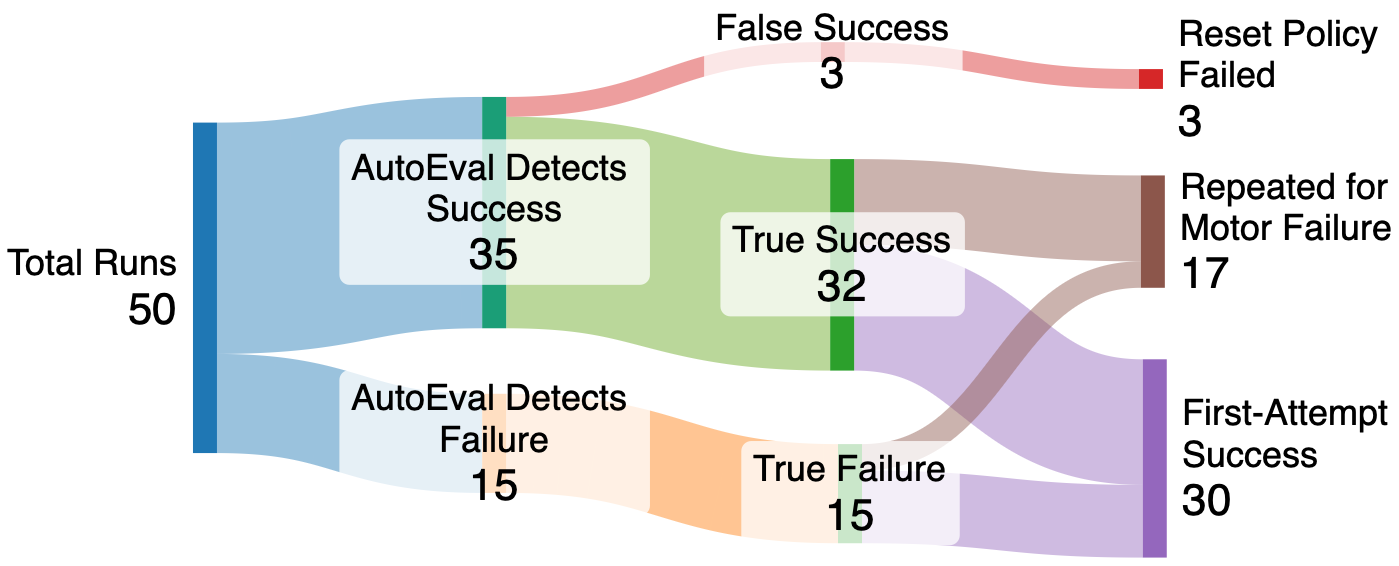}
    \caption{Analyzing $50$ \name{} runs on the \texttt{sink} scene:  the main failure modes is false positive results because the reset policy failed to reset the scene.}
    \label{fig:failure-modes}
    \vspace{-1em}
\end{figure}

While our previous experiments show that \name{} closely matches human-run evaluations, we observe that over extended periods of operation errors occur occasionally. To better understand the sources of these errors and help the design of future autonomous evaluation cells, we perform a detailed analysis of all failures occurring in a $50$ episodes \name{} run on the ``put eggplant in blue sink'' task with the Open-$\pi_0$ policy. We visualize the outcome of our analysis in \cref{fig:failure-modes}. While many episodes experienced motor failure because of harsh contact with the scene, \name{} handles such failure automatically by re-running those trials, and only report evaluation trials that do not contain motor failures. We find that for only three out of 50~trials, the autonomous evaluation fails, since the episodes mistakenly get classified as successes and the reset policy fails. 

One key takeaway from this failure analysis is that our Bridge-\name{} setup is already very reliable with few errors, and that most room for improvement is in improving the \emph{efficiency} by reducing the number of motor failures during evaluation, e.g. by implementing a more compliant robot controller that prevents harsh environment interactions.

\vspace{-0.5em}
\section{Conclusion} 
\label{sec:conclusion}

In this work, we introduced \name, a system for autonomous evaluation of generalist robot policies in the real world. We demonstrated that \name{} can perform high-quality evaluations around the clock and with minimal human involvements across a range of commonly used robot evaluation tasks. Our experiments shows that \name{} evaluation results closely matches those of human-run evaluations, and are both more reliable and applicable to a wider range of tasks than prior simulation-based evaluation approaches. In an effort to make real-robot evaluation widely available and more comparable, we provide public access to two \name{} evaluation cells for popular BridgeData~V2 evaluation tasks, for which users can submit their policies online for evaluation, and receive detailed evaluation reports. We hope that this work will inspire more \name{} evaluation cells to be set up across institutions to form a diverse automated evaluation framework, which will significantly speed up robot learning research.

\vspace{-0.5em}
\section{Limitations} 
\label{sec:limitations}

\noindent \textbf{\name{} environment creation time.} Our current approach for creating new environments for automated evaluation requires some up-front human effort to train the reset policy and success classifier. In our experience, the complete process only takes a few hours for a new scene and is quickly outweighed by the time savings of autonomous evaluation, but future work can explore more efficient ways of constructing reset policies and success classifiers to further reduce the effort for setting up a new scene for autonomous evaluation. We also expect that future improvements to vision foundation models and generalist policies will make the training of robust success classifiers and reset policies easier, possibly to the point where we can ``train'' these modules simply by providing a handful of examples in context. %

\noindent \textbf{Evaluating policy robustness.} There are various dimensions of out-of-distribution robustness we may be interested in evaluating for robot policies, e.g., robustness to varying camera angles, distractor objects, lighting conditions, or table textures (see \citet{gao2025taxonomy} for a more comprehensive taxonomy). Varying each of these axes in a controlled way as part of an \emph{automated} evaluation pipeline may require major engineering efforts, and \name{} currently does not support such evaluations. In the future, a decentralized network of \name{} cells may be able to increase evaluation diversity across many of these axes.

\noindent \textbf{Mobile manipulation tasks.} Our experiments capture a set of robot manipulation tasks that are reflective of the kind of tasks commonly used for evaluating generalist robot policies today, where the primary focus is on table-top manipulation tasks. We believe that our approach will transfer well to a wide range of other single-arm and bi-manual table top manipulation tasks. However, mobile robot tasks, particularly mobile manipulation tasks, may pose new challenges e.g. with regards to robust resets at room scale, success estimation under partial observability, and operational safety, all of which pose important directions for future work. 

\noindent \textbf{Binary success metrics.} \name{} evaluation currently only supports binary success estimates (did the policy succeed at a task or fail). When humans run evaluations, they can provide a more granular assessment of the policy's performance, including task progress scores and a qualitative analysis of the policy's proficiency. While \name{} users can obtain similar assessments from re-watching the logged evaluation videos, this is a time-consuming process. In future work, it would be exciting to investigate whether more granular performance analysis can be provided in an automatic evaluation framework, e.g., by querying powerful video summarization models.

\section*{Acknowledgments}
We would like to thank Kyle Stachowicz and Mitsuhiko Nakamoto for valuable discussions and feedback on an earlier version of the system. This work was partly supported by ONR N00014-25-1-2060 and NSF IIS-2150826. Pranav is supported by the NSF Graduate Research Fellowship.

\section*{Author Contributions}
Zhiyuan designed and implemented the \name{} system and Web UI and led the experiments. Pranav assisted in setting up the \name{} stations and running experiments. You Liang implemented the comparison experiments with SIMPLER and baseline methods and contributed to the \name{} station setup. Together, all three iterated on reset policy and success detector training. Karl and Sergey provided valuable guidance throughout the project and contributed to the paper writing.

\bibliographystyle{plainnat}
\bibliography{references}

\appendix
\renewcommand{\thesection}{\Alph{section}}

\section{Safety During Extended Autonomous Robot Operations}
\label{apdx:safety}

To ensure that the robots can autonomously and safely operate for a long time, we take several measures to ensure the safety of the robot and to preserve the scene. First, we set safety boundaries for the robot such that the policy cannot go beyond certain xyz axis (e.g. beyond the view of the camera) so that it does not run into objects unintentionally. 
Second, since the WidowX robot arms do not natively support impedance control, we limit the maximum effort on each of the robot joints, so that ineffective policies do not press too hard against objects and cause motor failure or damage the scene.  The common robot failure is due to joint failure when interacting and colliding with the objects in the scene, hence we constantly monitor and log the joint effort values, software reboot the joints at a safe arm position when joint errors are detected during each trial.
Third, we use the safety detectors described in Section~\ref{sec:method} to monitor and out-of-distribution and unexpected scenarios. 
Finally, we further ensure safety of the scene by taping the drawer and cloth to the table to prevent them from falling off the table, and add a thin plastic wrap over the yellow sink to prevent robot gripper getting jammed and damaged.

\section{Visualizations of \name{} Rollouts}
\label{apdx:filmstrip-visualizations}
\begin{figure*}[h]
    \centering
    \includegraphics[width=\linewidth]{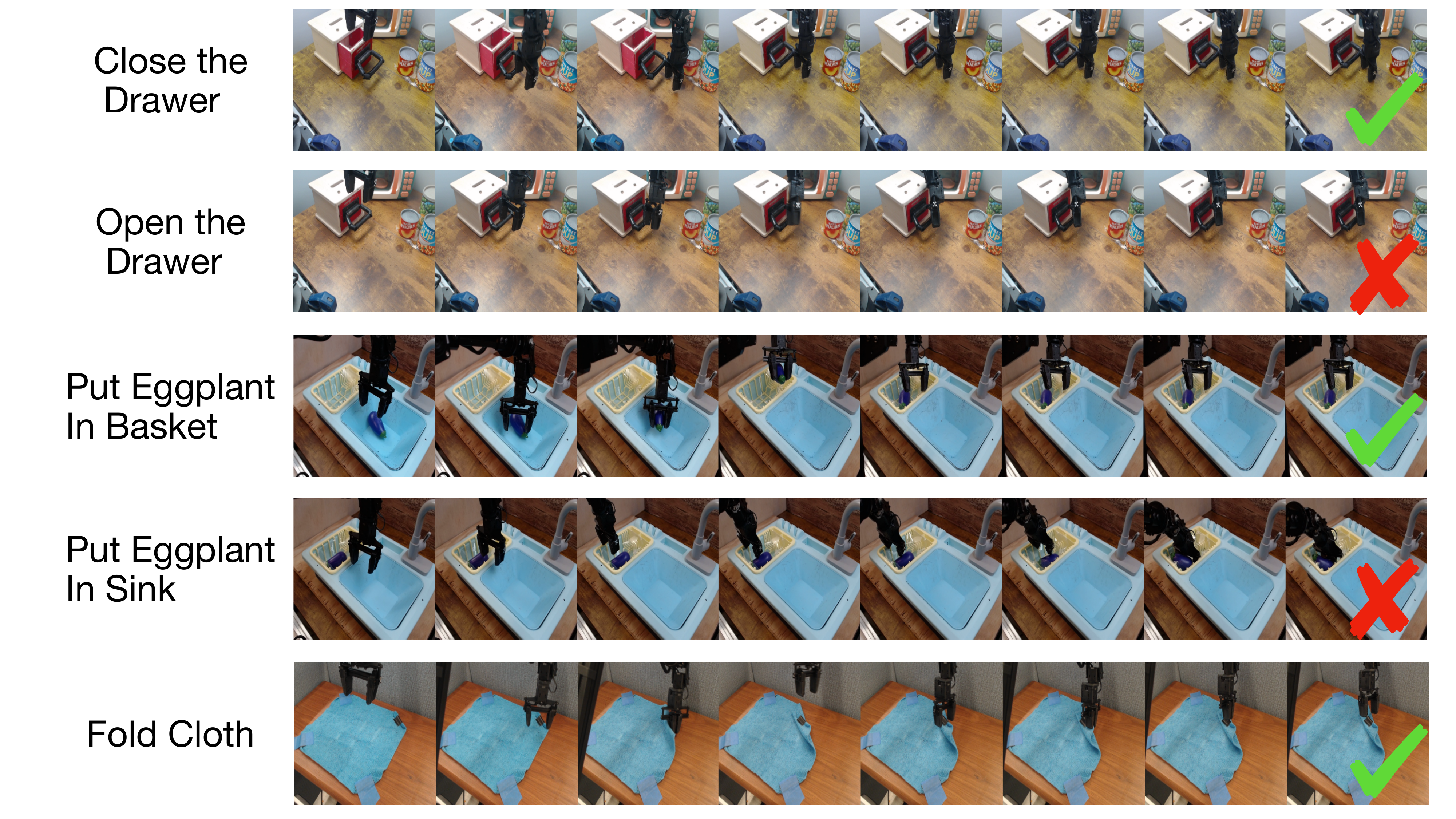}
    \caption{Samples of autonomous policy evaluation trials with \name{} on the five tasks. The classifier result of each task is visualized on the right hand side, and the reset policy is not shown.}
    \label{fig:film-strip} 
\end{figure*}

Figure~\ref{fig:film-strip} presents evaluation trajectories in the five different Bridge-\name{} tasks. The actual language commands fed to the evaluated policies are:
\begin{enumerate}
    \item \textit{``Close the drawer''}
    \item \textit{``Open the drawer''}
    \item \textit{``Put the eggplant in the yellow basket''}
    \item \textit{``Put the eggplant in the blue sink''}
    \item \textit{``fold the cloth from top right to bottom left''}
\end{enumerate}

\section{Detailed Evaluation Results on Bridge-\name}
\label{apdx:raw-eval-results}

In \cref{tab:autoeval-results} to \cref{tab:detailed-results}, we provide detailed evaluation results for our comparison of different scalable evaluation approaches across the five Bridge~V2 evaluation tasks.
For both tasks on the \texttt{Drawer} scene, we evaluate all policies for $70$ steps at maximum; for both tasks on the \texttt{Sink} scene, we run $100$ steps; for the \texttt{Cloth} scene, we run $80$ steps.

\begin{table}[h]
    \centering
    \small
    \setlength{\tabcolsep}{3pt}
    \begin{tabular}{|c|c|c|c|c|c|}
    \hline
    \textbf{Policy} & \multicolumn{2}{c|}{\textbf{Drawer}} & \multicolumn{2}{c|}{\textbf{Sink}} & \\
    \cline{2-5}
    & \textbf{Open} & \textbf{Close} & \textbf{To} & \textbf{To}  &\textbf{Fold}\\
    & \textbf{Drawer} & \textbf{Drawer} & \textbf{Basket} & \textbf{Sink}  &\textbf{Cloth}\\
    \hline
    OpenVLA   & 40/50 & 46/50 & 1/50  & 0/50  & 13/50 \\
    Open $\pi_0$  & 29/50 & 46/50 & 7/50  & 47/50 & 12/50 \\
    Octo      & 1/50  & 5/50  & 0/50  & 0/50  & 4/50  \\
    SuSIE-LL  & 0/50  & 1/50  & 0/50  & 0/50  & 0/50  \\
    SuSIE     & 1/50  & 18/50 & 0/50  & 0/50  & 9/50  \\
    MiniVLA   & 33/50 & 49/50 & 38/50  & 0/50  & 8/50  \\
    \hline
    \end{tabular}
    \caption{AutoEval results on five Bridge-\name{} tasks across six different generalist policies.}
    \label{tab:autoeval-results}
\end{table}

\begin{table}[h]
    \centering
    \small
    \setlength{\tabcolsep}{3pt}
    \begin{tabular}{|c|c|c|c|c|c|}
        \hline
        \textbf{Policy} & \multicolumn{2}{c|}{\textbf{Drawer}} & \multicolumn{2}{c|}{\textbf{Sink}} & \\
        \cline{2-5}
        & \textbf{Open} & \textbf{Close} & \textbf{To} & \textbf{To}  &\textbf{Fold}\\
        & \textbf{Drawer} & \textbf{Drawer} & \textbf{Basket} & \textbf{Sink}  &\textbf{Cloth}\\
        \hline
        OpenVLA   & 40/50 & 46/50 &  1/50 &  0/50 & 12/50 \\
        Open $\pi_0$  & 24/50 & 45/50 &  7/50 & 47/50 &  3/50 \\
        Octo      &  0/50 &  0/50 &  0/50 &  0/50 &  2/50 \\
        SuSIE-LL  &  0/50 &  0/50 &  0/50 &  0/50 &  0/50 \\
        SuSIE     &  2/50 & 13/50 &  0/50 &  0/50 & 10/50 \\
        MiniVLA   & 32/50 & 49/50 &  38/50 &  0/50 &  8/50 \\
        \hline
    \end{tabular}
    \caption{Ground truth human evaluation results for the five Bridge-\name{} tasks across six different generalist policies.}
    \label{tab:detailed-results}
\end{table}

\section{Evaluation  on Bridge-SIMPLER~\citep{li24simpler}}
\label{apdx:raw-simpler_eval-results}

In \name{} , we introduced a new Drawer Scene to the existing SIMPLER~\citep{li24simpler} setup for the WidowX robot. The scene was visually matched with the \name{}'s Drawer setup, and overlaid with the same background to ensure consistency. A 3D model of the drawer, with exact dimensions matching the real-world setup, was also created. This scene introduced two evaluation tasks: \textit{"open and close the drawer"}. To add variability to each evaluation trial, we randomized the end effector's initial pose, the drawer's initial pose, and the lighting conditions in the background.

In addition to the Drawer Scene, \name{} includes a Sink Setup, which closely resembles the existing SIMPLER~\citep{li24simpler} Sink Scene. In SIMPLER, the task here is the \textit{"move the eggplant to the basket"} task. We also introduced a reverse task, \textit{"move eggplant to the sink,"} effectively making the scene reset-free. This allows for both forward and reverse tasks in the same environment.

With these two scenes and four tasks, we conducted 50 runs for each scene across five different generalist policies. The detailed results are shown in \cref{tab:simpler-results} 

\begin{table}[h]
    \centering
    \small
    \begin{tabular}{|c|c|c|c|c|}
        \hline
        \textbf{Policy} & \multicolumn{2}{c|}{\textbf{Drawer Scene}} & \multicolumn{2}{c|}{\textbf{Sink Scene}} \\
        \cline{2-5}
        & \textbf{Open} & \textbf{Close} & \textbf{To} & \textbf{To} \\
        & \textbf{Drawer} & \textbf{Drawer} & \textbf{Basket} & \textbf{Sink} \\
        \hline
        OpenVLA & 32/50 & 2/50 & 1/50 & 0/50 \\
        Open $\pi_0$ & 34/50 & 24/50& 45/50& 6/50 \\
        Octo & 3/50 & 0/50 & 6/50 & 3/50 \\
        SuSIE-LL & 1/50 & 0/50 & 0/50 & 0/50 \\
        SUSIE & 0/50 & 41/50 & 7/50 & 0/50 \\
        MiniVLA & 30/50 & 23/50 & 10/50 & 2/50 \\
        \hline
    \end{tabular}
    \caption{Evaluation Results on SIMPLER~\citep{li24simpler} for Drawer and Sink Scene on four tasks and six different policies.}
    \label{tab:simpler-results}
\end{table}

\section{Computing Action Validation MSE between policies}
\label{sec:action_mse_policies}

We sample 400 trajectories from the validation set of BridgeData~\citep{walke2023bridgedata} to compute the action mean squared error (MSE) for each policy. The results are shown in \cref{tab:validation-mse-results}. Consistent with the findings in SIMPLER~\citep{li24simpler}, this illustrates a weak correlation of task success rate on AutoEval with validation MSE.

\begin{table}[h]
    \centering
    \footnotesize
    \begin{tabular}{|l|c|c|c|c|}
        \hline
        \textbf{Policy} & \multicolumn{2}{c|}{\textbf{200 Trajectories}} & \multicolumn{2}{c|}{\textbf{400 Trajectories}} \\
        \cline{2-5}
        & \textbf{MSE} & \textbf{Norm MSE} & \textbf{MSE} & \textbf{Norm MSE} \\
        \hline
        OpenVLA & 0.0143 & 1.362 & 0.015 & 1.431 \\
        Open $\pi_0$ & 0.082 & 1.433 & 0.085 & 1.495 \\
        Octo & 0.0214 & 1.504 & 0.023 & 1.611 \\
        GCBC & 0.008 & 0.817 & 0.009 & 0.870 \\
        SUSIE & 0.018 & 1.1579 & 0.018 & 1.244 \\
        \hline
    \end{tabular}
    \caption{Average Validation MSE across Policies on $400$ random trajectories from BridgeV2 Dataset. Norm MSE represents the MSE of normalized actions, while MSE represents the MSE of raw action magnitudes.}
    \label{tab:validation-mse-results}
\end{table}

\section{Success Classifier in Bridge-\name{} cells}
\label{apdx:success-classifier}

To train success classifiers for the Bridge-\name scenes, we finetune the Paligemma VLM to act as a classifier. We manually collect a dataset of roughly 1000 images for each scene, and manually label them. We form VQA questions with the labels, and finetune the base 3B parameter VLM with quantized LoRA using a learning rate of $2e-5$, batch size of $4$ for  $80$  iterations. For some scenes, we combines the success classifier and the safety detector into a single fine-tuned VLM: we train the VLM to output ``invalid'' (in addition to classifying success) when there are out-of-distribution cases that prevent evaluation from proceeding autonomously (\eg object out of reach).

For each evaluation scene, approximately 1000 image frames are collected to fine-tune the VLM. The corresponding language prompts are:
\begin{enumerate}
    \item Sink Scene: \textit{"is the eggplant in the sink or in the basket? answer sink or basket or invalid"}
    \item Drawer Scene: \textit{"is the drawer open? answer yes or no"}
    \item Cloth Tabletop Scene: \textit{"is the blue cloth folded or unfolded? answer yes or no"}
\end{enumerate}

We evaluate our classifier both by running it on a held-out test set of roughly $100$ images and by teleoperating the robot and running the classifier on all the image observations throughout the trajectory. We choose to deploy success classifiers in AutoEval that have an accuracy of $>95\%$. When the classifier trained on the initial $\sim 1000$ images does not achieve this accuracy threshold, we found it helpful to improve classifier performance by rolling out the trained model, identifying incorrect predictions and collecting these images, and retraining on these ``hard'' examples.

\section{Reset Policy in Bridge-\name{} cells}
\label{apdx:finetune-reset-policy}

To train reset policies for the Bridge-\name cells, we finetune the generalist OpenVLA policy with LoRA, with batch size 64 and learning rate $10^{-4}$ for $1000$ iterations. 
For each scene, we collect $50-100$ demonstration trajectories via teleoperation, and train with a standard behavior cloning loss. We also use a scripted policy for one of our reset policy - "Close the Drawer" task, where the reset success rate is not sensitive to variation in scene.

Similar to the success classifiers, we choose to deploy reset policies that have a success rate of $>95\%$.

\section{Step-by-step \name{} Construction Guide}
\label{sec:appx_step_by_step_guide}

Below, we provide a step-by-step guide for creating an \name{} setup for a new evaluation tasks. Refer to our code release at \url{https://github.com/zhouzypaul/auto_eval} for code on each of the steps and detailed instructions on how to run the code. The full process takes approximately 3~hours of active human effort and a total of 5~hours including model training time for reset policy and success detector.

\begin{enumerate}
\item \textbf{Train Reset Policy}: 
Start by collecting approximately $50-100$ high-quality robot demonstrations of resetting behavior from sensible final states of policy rollouts. Try to cover a diverse set of ``reset start states'', including those that failed the original task, to obtain a robust reset policy. Once you collected the dataset, fine-tune a generalist policy like OpenVLA~\citep{kim2024openvla}, e.g., using LoRA fine-tuning, on your the small demonstration dataset. 
If you find that the reset is unreliable and fails often, consider collecting more reset demonstrations particularly on the positions where the reset policy fails and re-train the policy. 
For a small set of tasks that has more structure, you can also use \emph{scripted} policies to reset the scene. See our code release for code to record tele-operated policies for WidowX robots and replaying it to reset the scene.
An easy way to make reset policies stronger is to simply execute it for multiple times if it fails. Proceed when your reset has success rate of $>95\%$.
\item \textbf{Train Task Success Classifier (And Safety Detector)}:
While the success classifier and the safety detectors serve two different functions, in practice you can train a combined three-way (success, failure, invalid) classifier that acts as both the success and safety detector. This classifier will output ``invalid'' when OOD events happen (e.g. objects out of reach) and human intervention is needed, else it will output whether the task is successfully completed or not.
Collect approximately 1000~images of success and failure (and invalid) states. Be sure to collect lots of failures (and invalid) states because there are many ways in which the robot can fail. 
Then fine-tune a vision-language model like Paligemma~\citep{beyer2024paligemma} on this dataset. 
Test the performance of your classifier by tele-operating the robot and scoring the observations along the trajectory. You can improve the classifier by saving the observations that it mis-classifies and re-train by incorporating these ``hard examples'' into the original dataset of images. Proceed when your success classifier has accuracy $>95\%$.
\item \textbf{Set Up Safety and Robustness Measures}:
We have implemented multiple safety measures described in \cref{apdx:safety} for the WidowX robot. To set up new \name{} tasks on WidowX robots (or ViperX or similar robots), you can directly use our infrastructure; to set up \name{} on a new robot embodiment, consider implementing the following safety measures.
First, if your robot does not have an integrated p-stop that prevents forceful collisions with the environment, implement a limit on the motor current to prevent high-force contact with the environment that may damage the robot and the scene. Also implement a software mechanism to reboot the motors when they fail.
Second, use a workspace boundary to limit the reach of the robot: limit the robot from reaching out-of-scene objects and prevent the robot from removing objects that are in the scene.
Finally, use an ``on-call'' system that sends push notifications to human monitors when the robot reports an irrecoverable safety issue or the reset policy fails for $N\approx3$ times in a row (as determined by the success detector). We implement a Slack bot that sends notifications through Slack channels.

\item \textbf{Prepare for Policy Submission}: 
Power on the robots and start the low-level robot controllers. We set up the robot environment as a server that waits to receive actions from the policy.
Then, start the webserver to access the UI for tracking the evaluation jobs queue and submitting a policy through the webapp.
Next, host your policy that needs to be evaluated as a server.
Finally, submit the IP and port of your policy server to the \name{} web UI as an evaluation job to the \name{} system. The evaluation job will be automatically queued and ran.
See the code release for more detailed instructions.

\end{enumerate}

\section{Bridge-\name{} Deployment Details}
\label{apdx:deployment-details}

As described in \cref{sec:bridge-autoeval-details}, we open access to our Bridge-\name{} cells to the research community. The two different \name{} cells accepts and executes jobs in parallel.
While the two WidowX robots will accept evaluation jobs 24/7, we enforce a $20$ minute rest period every $6$ hours where the robot will torque off and let the motors cool off (see \cref{fig:consistent-across-time} for why this is necessary). The reset period will only happen between evaluation jobs.

Since we host the reset policies for the four tasks in Bridge-\name{} 24/7, we optimize for lightweight policies (as compared to the fine-tuned OpenVLA reset policy we use in \cref{sec:results}). For the two tasks on \texttt{Drawer}, we use scripted reset poliy; for the two tasks on \texttt{Sink}, we fine-tune MiniVLA~\citep{belkhale2024minivla} on the same demos. We find that all reset policies have success rate $>95\%$.

\section{Evaluation Results Reproducible Across Months}
\label{apdx:two-months-reproducible}

We find that \name{} reproduces results even after more than 2 months of continued use, demonstrating its robustness to aging effects.
We compare AutoEval results that are two months apart for two policies on three tasks as shown in \cref{tab:two-months}. During the two months, AutoEval operated continuously for a rough total of $200$ hours. \cref{tab:two-months} shows that all evaluations perform similarly when evaluated two months apart, and the reset policy and success classifiers still have accuracies \textbf{96\%} and \textbf{96\%} respectively. We attribute such robustness to (1) safety controllers (Appendix A) limiting robot joint efforts to prevent high-force contact and damages, and (2) foundation model pre-training (Paligemma VLM, OpenVLA) making policies and detectors resilient to minor scene changes. Over two months, we have observed minimal “aging” -- e.g. there are light scratches on the drawer upon close examination, but they are not visible in the 256x256 pixel policy image observations and does not impact the drawer physics. 

\begin{table}[h]
    \centering
    \small
    \setlength{\tabcolsep}{5pt}
    \begin{tabular}{|c|c|c|c|}
    \hline
    \textbf{Policy} & \textbf{Open} & \textbf{Close} & \textbf{Eggplant}\\ %
    & \textbf{Drawer} & \textbf{Drawer} & \textbf{to Basket} \\ %
    \hline
    OpenVLA (old)   & 39/50 & 48/50 & 4/50  \\%& 1/50  \\
    OpenVLA   & 40/50 & 46/50 & 1/50 \\% & 0/50  \\
    \hline
    \textbf{Avg. $\Delta$ Success} & \textcolor{black}{+2\%} & \textcolor{black}{-2\%} & \textcolor{black}{-6\%}\\% & -2\% \\
    \hline
    \hline
    Open $\pi_0$ (old)  & 30/50 & 45/50 & 9/50 \\% & 35/50 \\
    Open $\pi_0$  & 29/50 & 46/50 & 7/50 \\% & 47/50 \\
    \hline
    \textbf{Avg. $\Delta$ Success} & \textcolor{black}{-2\%} & \textcolor{black}{+2\%} & \textcolor{black}{-4\%}\\% & +24\% \\
    \hline
    \end{tabular}
    \caption{AutoEval results obtained two months apart: results remain highly consistent across two different policies on three different tasks. All correlate well to human evaluations.}
    \label{tab:two-months}
    \vspace{-1.5em}
\end{table}

\section{Initial States in Bridge-\name{} Cells}
\begin{figure}[h]
    \centering
    \includegraphics[width=0.31\linewidth]{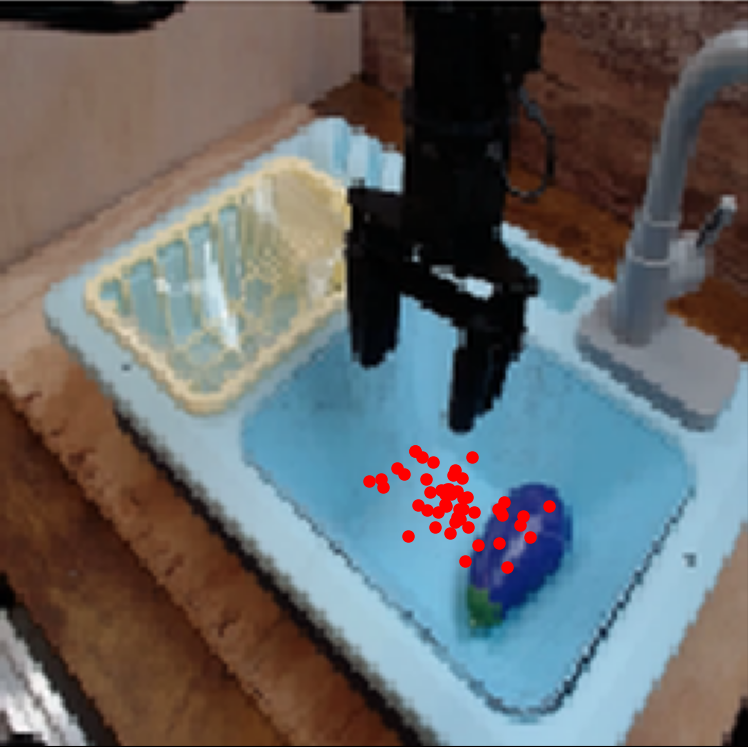}
    \includegraphics[width=0.31\linewidth]{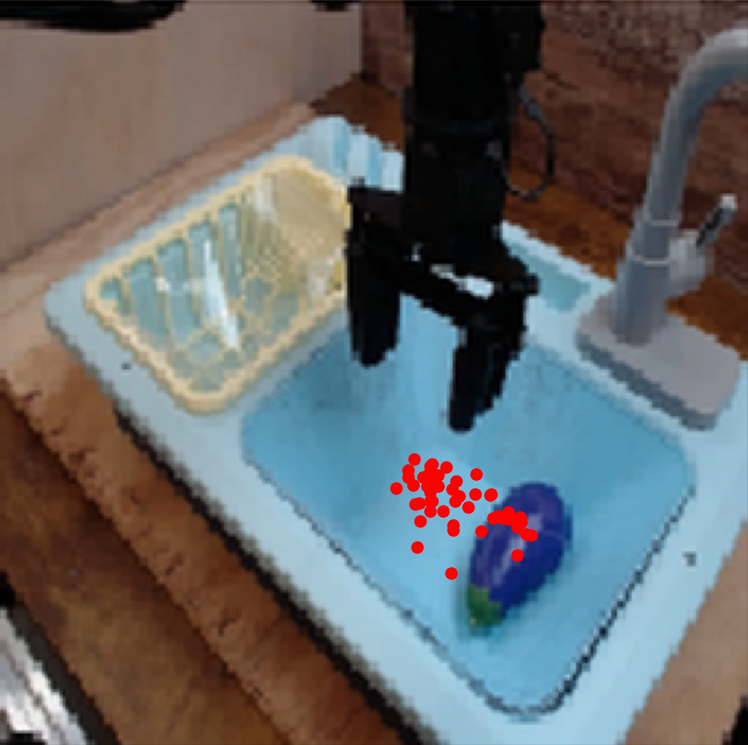}
    \includegraphics[width=0.31\linewidth]{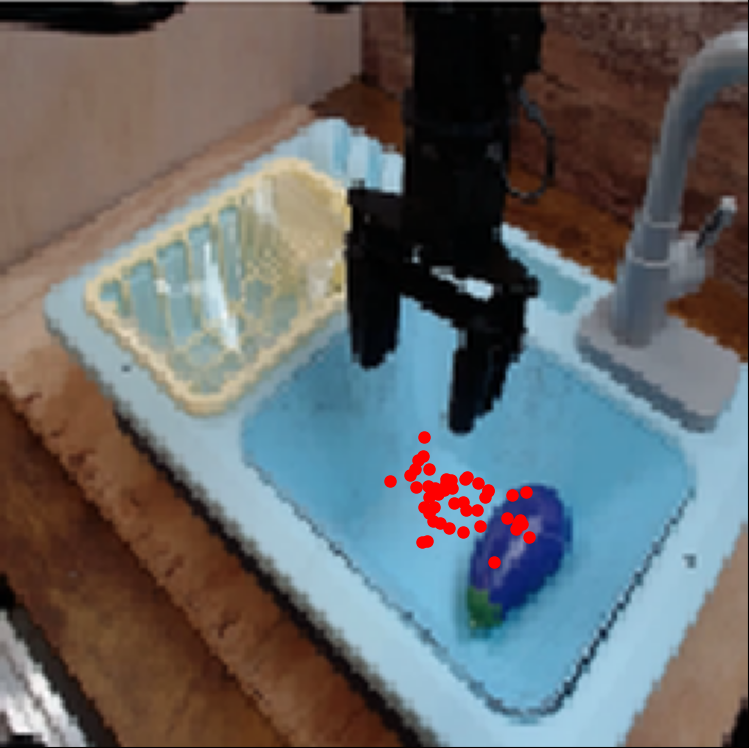}
    \caption{Initial state distribution for $3$ different AutoEval runs is consistent. Red dots show the centroid position of the eggplant. Each run uses $50$ trials.}
    \label{fig:init-state}
\end{figure}

We find that our learned reset policy is able to reset to a consistent distribution of initial states. 
As an example, we plot the centroids of all eggplant initial positions for three representative AutoEval runs of the ``Eggplant to Basket'' task in \cref{fig:init-state} (50 trials each). Qualitatively, we find that the reset distributions of other tasks are similarly overlapping, and also roughly cover the task distribution.

\section{Improving \name{} with Additional Human Involvement}

Though \name{} results highly correlate with ground truth human-run evaluations, it is not perfect (as shown in \cref{fig:failure-modes}). Additional human effort, when available, can further improve \name{}'s accuracy.
The easiest and most effective way to apply extra human effort can be spent going through the evaluation report after AutoEval finishes to remove the runs where the reset policy fails, and relabel the success manually. Going through $50$ trials of \name{} roughly takes $1-2$ minutes of human time. This enables ground-truth judgment of evaluations runs while still saving the majority of human time required in robot evaluations.

\end{document}